\documentclass[conference]{IEEEtran}
\IEEEoverridecommandlockouts

\usepackage{cite}
\usepackage{amsmath,amssymb,amsthm}
\usepackage{mathtools}
\usepackage{enumitem}
\usepackage[table]{xcolor}
\usepackage{algorithm}
\usepackage{algorithmic}
\usepackage{graphicx}
\usepackage{multirow}
\usepackage{booktabs}
\usepackage{tabularx}
\usepackage{caption}
\usepackage{hyperref}
\usepackage{url}
\graphicspath{{figures/}}

\providecommand{\R}{\mathbb{R}}

\providecommand{\R}{\mathbb{R}}

\newcommand{\NF}{\mathrm{NF}}

\begin{document}

\title{Diagnosing Faults in Reinforcement Learning Simulators with Canonical Polynomial Invariants}

\author{
    \IEEEauthorblockN{
        Tesfay Zemuy Gebrekidan \\
        \textit{Independent Researcher} \\
        \textit{Southampton, United Kingdom} \\
        tzemuy13@gmail.com
    }
    \and
    \IEEEauthorblockN{
        Hadush Hailu Gebrerufael \\
        \textit{Computer Science, MIU} \\
        \textit{Iowa, USA} \\
        hadush.gebrerufael@miu.edu
    }
}

\maketitle

\begin{abstract}
A large literature incorporates physical structure into learned dynamics on the premise that models respecting the underlying physics should predict better. We test this premise using exact polynomial invariants recovered from trajectories and represented canonically as reduced Gr\"obner bases over $\mathbb{Q}$. We find that exact polynomial invariants provide little additional benefit for residual-based downstream uses. On Gymnasium's Acrobot, consistency regularisation reduces algebraic residuals without improving long-horizon rollout fidelity, while one-step validation error is more strongly associated with long-horizon accuracy; similarly, shaping potentials recovered from systems with mass errors of up to $100\%$ achieve similar learning performance to the physically correct potential. The main benefit observed here instead appears in diagnosis, where canonical representations permit exact ideal-equality and membership decisions, and where a reference generating set supplies a physical vocabulary that generic two-sample tests do not have: handed the same per-generator residuals, a Kolmogorov--Smirnov baseline localises almost as well, so the advantage comes from the generating set and not from the algebra that reads it. We develop a two-level diagnostic comprising screening, which identifies the violated physical constraint, and attribution, which recovers the faulty invariant and identifies the responsible physical parameter. Normal-form deflation removes algebraically trivial multiples, while quotient-space recovery avoids tolerance-based nullspace selection. Across fifteen injected faults, screening localises all fifteen when no observation noise is added, with no observed false alarms. Attribution identifies the responsible parameter in all seven single-parameter faults. Paired difference tests detect all faults but provide no constraint-level localisation, showing that the advantage is localisation and not detection. Finally, applying the diagnostic to 350 release pairs across eleven RL environments finds no evidence of changed simulator dynamics, but exposes a stale forward-kinematics read in Reacher's observation function, an integrator-drift ceiling on Acrobot, and a chaotic-amplification limit on comparisons in InvertedDoublePendulum. In conclusion, exactness is more useful for diagnosing violations of physical constraints than for improving prediction.
\end{abstract}


\section{Introduction}
\label{sec:intro}

Building physical structure into learned dynamics is an active research
direction. Conservation laws are added to training losses, Hamiltonian and
Lagrangian structure is built into model architectures, and kinematic
constraints are enforced by projection. These methods assume that a model
respecting the physics predicts better. This assumption is rarely tested
against the strongest available form of that structure, because the invariants
these methods use are learned or numerically approximate rather than exact. In
this work, we use invariants that are exact and canonical, and evaluate what
this exactness provides. We find that it provides little benefit for
prediction, but is valuable for diagnosis.

This is because exactness and canonicity provide a property that approximate
invariants do not: decidable equality. Two ideals are equal if and only if
their reduced bases are equal, and ideal membership is decided by normal-form
reduction alone \cite[Chapter 2, Sections 6 and 7]{cox2015ideals}. Neither
property holds for a learned latent function or an approximate border basis.
Most downstream uses of an invariant only need a differentiable residual, for
which an approximate invariant performs about as well as an exact one. The
uses that need the canonical form are those that must decide something:
whether two systems have the same algebra, and if not, where they differ.
Canonical form applies only to the vanishing-constraint ideals, recovered as
reduced Gr\"obner bases; conserved quantities such as energy form a
subalgebra with no finite canonical basis in general, and are instead
recovered up to gauge and deduplicated by functional independence
(Section~\ref{sec:prelim}).

These decisions matter in practice because reinforcement learning (RL) relies
on simulators, and simulators are software: they can contain mistyped
constants, terms dropped from hand-derived equations of motion, integrators
swapped for speed, or observation vectors assembled in the wrong order.
Existing tools for catching such errors are aggregate measures: a validation
error, a rollout divergence curve, or at best a two-sample test between logged
state distributions. Each of these returns a single number, which cannot say
what is wrong. A practitioner who learns that their model is inaccurate does
not learn, for example, that the second link's mass is $1.3$ and not $1.0$. We
address this with a two-level diagnostic. \emph{Screening} tests each
reference generator's residual on the system under test and identifies the
constraint that is violated. \emph{Attribution} then recovers the invariant,
fits a parametric template to it, and returns the responsible constant.

Our contributions are fourfold. First, we introduce a two-level algebraic diagnostic for RL simulators based on polynomial invariants. The diagnostic separates screening, which identifies the violated physical constraint, from attribution, which recovers the faulty invariant and identifies the responsible physical parameter. Second, we characterise what exact canonicalisation provides and what it does not. Canonical forms make equality of invariant ideals decidable, enabling exact comparison of recovered and reference physics. We show, however, that exactness is unnecessary for screening, where substantially perturbed generators localise faults equally well, but is essential for ideal-equality decisions. A control that hands the same per-generator residuals to a generic two-sample baseline makes the division explicit: the baseline then localises almost as well, so the advantage in Section~\ref{sec:e1} comes from the reference generating set and not from the particular statistic that reads it. Third, we develop two algorithmic components that make the diagnostic
practical. Normal-form deflation removes algebraically redundant multiples
exactly, and quotient-space recovery avoids tolerance-based nullspace
dimension estimation, enabling invariant recovery without heuristic nullspace
selection. Fourth, we apply the diagnostic to 350 release pairs across eleven RL environments. We find no evidence of changed simulator dynamics, but instead expose implementation properties including integrator sensitivity, an observation inconsistency in Reacher, and the limits imposed by chaotic dynamics on long-horizon release comparisons.

\section{Related Work}
\label{sec:related}

\paragraph{Physical structure and its discovery.} In many existing works, conservation laws and
kinematic constraints are added to learned models as differentiable penalties
\cite{beucler2021enforcing,djeumou2022neural}, built into Hamiltonian and
Lagrangian architectures \cite{greydanus2019hamiltonian,cranmer2020lagrangian},
or discovered contrastively and enforced by projection
\cite{zhang2023concernet}. The invariants themselves come either from symbolic
regression and conservation-law discovery
\cite{udrescu2020aifeynman,brunton2016sindy,kaiser2018discovering,liu2021aipoincare}
or from
approximate vanishing-ideal methods that build ideals from noisy samples
\cite{heldt2009approximate,livni2013vca,wirth2023avi}. All of these produce a residual
and return floating-point objects with no canonical form, so none supports a
test for equality between two discovered invariant sets, the property the
diagnostic needs. Section~\ref{sec:e2} tests whether the exact version
outperforms the approximate one at prediction, and finds that it does not.

\paragraph{Simulator fidelity, and what the field does about it.} The gap
between a simulator and the system it stands for is usually managed by
randomising over it \cite{tobin2017domainrand,peng2018simtoreal} and not by
locating the discrepancy, and reproducibility studies document how often
RL results move for reasons nobody can localise
\cite{henderson2018deeprl}, implementation-level choices sometimes accounting
for more of a published gain than the algorithm they are credited to
\cite{engstrom2020implementation}. Version drift between shipped releases
\cite{towers2024gymnasium} is a recognised instance. Metamorphic testing
\cite{chen2018metamorphic} checks software using relations that must hold
between inputs and outputs instead of known correct answers; the screen does
the same thing structurally, except that there the relations are hand-written
and here they are discovered and canonical. Dynamic invariant detection
\cite{ernst2007daikon} and property differencing \cite{yang2014propertydiff}
compare program versions through the properties each satisfies, but over
program variables with template-enumerated relations and an attached
confidence value; our method instead returns a physical constraint named from
a canonical basis over $\mathbb{Q}$, together with the responsible constant,
and our search found no application of either line of work to the physics of
a simulator release. The supplementary material (Appendix~A) discusses both in detail.

\paragraph{System identification and safety verification.} \emph{System
identification}
\cite{ljung1999sysid} presumes the model structure is right and estimates
within it, whereas the screen decides whether the structure is right at all and
attribution runs only afterwards, refusing when no parameter explains the data.
\emph{Safety verification} uses Gr\"obner and sum-of-squares certificates for
barrier functions \cite{prajna2004barrier,parrilo2003sos} over semialgebraic
sets, whereas our objects are equalities. The supplementary material (Appendix~A) discusses
both, together with the world-model literature Section~\ref{sec:e2} evaluates,
reward shaping \cite{ng1999shaping,burda2019rnd}, and Gr\"obner bases in
program-invariant generation.

\section{Setup: Ideals, Conserved Quantities, and Canonical Forms}
\label{sec:prelim}

The invariants used here come in two kinds. Constraints that vanish on every
admissible state are one, quantities that stay constant along a trajectory are
the other, and only the first is canonicalised by a reduced Gr\"obner basis. For admissible states
$\mathcal{M} \subseteq \R^n$ the vanishing set
\begin{equation}
\label{eq:vanishing-ideal}
I(\mathcal{M}) \;=\; \bigl\{\, p \in \R[x_1,\dots,x_n] : p(x) = 0\ \forall\, x \in \mathcal{M} \,\bigr\},
\end{equation}
is an \emph{ideal}. In control, such ideals are common and exact: rigid-link
constraints, the identity $\cos^2\theta + \sin^2\theta - 1$ that holds whenever
an angle reaches a policy as a $(\cos,\sin)$ pair, forward kinematics, loop
closures and stoichiometric conservation. For a fixed monomial order the reduced
Gr\"obner basis of such an ideal is unique \cite[Chapter 2, Section
7]{cox2015ideals}, and every constraint just listed holds \emph{regardless of
the control input}, so ideals are the right primary object for an actuated
system. Uniqueness is not minimality, however, since the reduced basis of the
twisted cubic has three elements for a two-generator ideal
(Remark~1 in the supplementary material).

Conserved quantities are not an ideal but a \emph{subalgebra}, because $E$ being
conserved implies that $E^2$ is conserved, so reduced Gr\"obner bases do not
canonicalise them. Energy on Acrobot is whatever the initial condition makes it,
so a search for vanishing polynomials pooled across trajectories recovers the
unit-norm constraints and misses the energy entirely. We search instead a
\emph{difference dictionary}
\begin{equation}
\label{eq:diff-dict}
\Delta_d \;=\; \bigl\{\, m(s') - m(s) : m \in \mathcal{P}_{\le d} \,\bigr\},
\end{equation}
whose nullspace vectors are
conserved at every energy level simultaneously, and we remove redundancy by
extracting a maximal functionally independent subset certified by Jacobian rank,
which discards $E^2$ given $E$. This certifies irredundancy, not canonicity; maximal independent subsets are not unique. For actuated damped systems the power balance replaces conservation, and has a drift floor of its own; the supplementary material (Appendix~B) develops all three objects.

\section{Deflation and Tolerance-Free Recovery}
\label{sec:scaling}

\begin{figure*}[t]
\begin{center}
\includegraphics[width=1.0\textwidth]{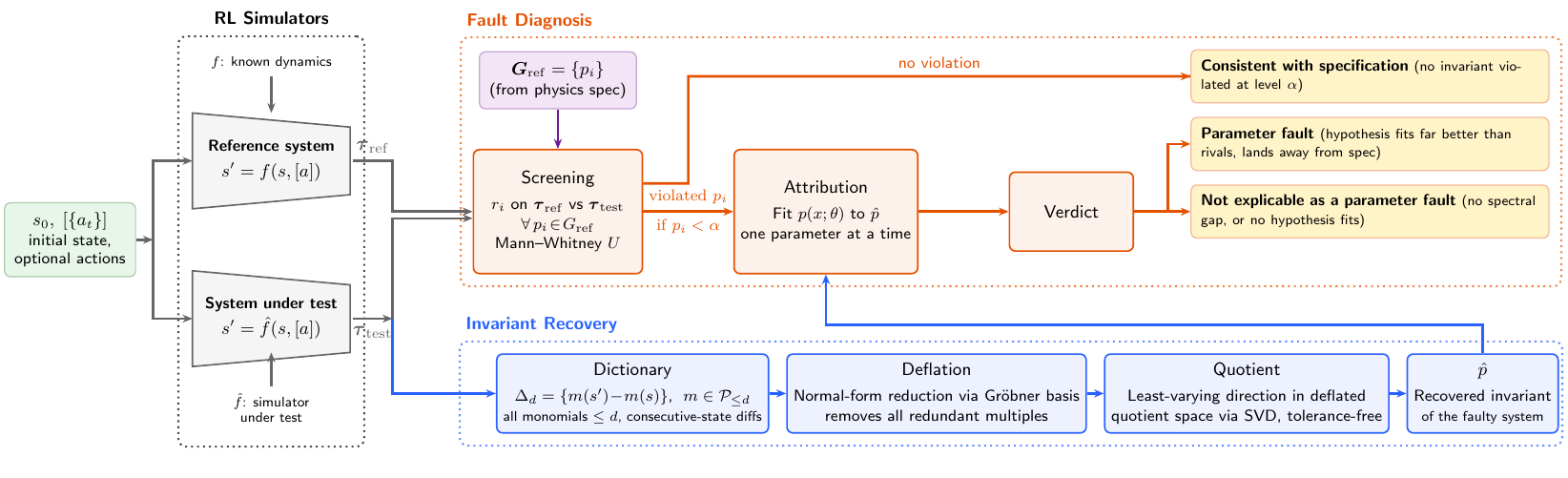}
\end{center}
\caption{The diagnostic. Invariant recovery (bottom) turns trajectories from
the system under test into a candidate invariant $\hat p$
(Sections~\ref{sec:scaling}--\ref{sec:quotient}). Screening (top left) tests
each reference generator's residual and reports which $p_i \in G_{\rm ref}$
is violated; attribution (top right) fits $\hat p$ against a parametric
template one parameter at a time and returns a verdict.}
\label{fig:pipeline}
\end{figure*}

Two obstacles stood between a monomial dictionary and a usable generator, and
both had to be removed before the diagnostic could exist.
The supplementary material (Appendix~C) covers dictionary sizes, support-search cost,
coordinate lifting and nondimensionalisation.

\subsection{Deflation by normal-form reduction}
\label{sec:deflation}
The degree-$d$ nullspace is dominated by algebraically trivial content: if $g$
is a known degree-$2$ invariant, every $m \cdot g$ with $\deg m \le d-2$ lies in
it and none of them is new information. Orthogonal projection removes $\mathrm{span}(g)$, one direction:
\begin{equation}
\label{eq:proj}
V \leftarrow \bigl(I - gg^\top/\|g\|^2\bigr)\,V,
\end{equation}
where the trivial content is the whole filtered piece
$\langle g\rangle \cap \mathcal{P}_d$, of dimension
$\binom{n + d - \deg g}{d - \deg g}$. Normal-form reduction against a reduced
Gr\"obner basis of $\langle G_{\rm known}\rangle$ removes that space exactly as algebra:
\begin{equation}
\label{eq:deflation}
V \leftarrow \NF_{G_{\rm known}}(V),
\end{equation}
so that all algebraically redundant multiples vanish in one step. On Acrobot, trivial multiples account for nearly the entire degree-$3$ nullspace. Table~S3 in the supplementary material reports the ablation alongside the
recovered direction's alignment with the true energy.

\subsection{Recovery in the quotient, without a tolerance}
\label{sec:quotient}
Attribution needs the invariant of the \emph{faulty} system, which has to be
recovered and not assumed. Estimating a numerical nullspace against a tolerance
and then deflating what was found fails under noise: with noisy observations
$c_1^2 + s_1^2 - 1$ no longer vanishes, so the fourteen trivial multiples of the known degree-2 invariant in the Acrobot degree-3 dictionary leave
the nullspace entirely, and the tolerance that would admit the energy admits a
great deal else first. Reversing the order removes the failure and the absolute
tolerance together. Let $\Pi$ be an orthonormal basis of the image of the
normal-form map $\NF_G$ on
$\mathcal{P}_d / (\langle G_{\rm known}\rangle \cap \mathcal{P}_d + \R)$,
computed once in exact rational arithmetic; we build the difference design
matrix $\Phi$, restrict it as $\Phi\Pi^\top$, and take the right singular vector
belonging to the smallest singular value. No nullspace dimension is estimated
from the data anywhere, so the units-dependent absolute tolerance is gone. What
remains is a dimensionless threshold on the spectral gap
$\sigma_{\min}/\sigma_{\text{next}}$, fixed at $0.1$ throughout and invariant to
a rescaling of the monomial dictionary columns and to the sample count; a system that
conserves nothing has no gap and is refused.

Deflation is a precondition here, not an efficiency measure. Without the
projection the least-varying direction lies entirely inside the trivial piece, at
a measured fraction of $1.0000$, and the fourteen-dimensional degeneracy there
means no single direction is identified at all; with it the recovered direction
aligns with the energy to $1.0000$. Snap-rounding is deliberately bypassed,
because a rationality prior is a prior on the system being \emph{correct}, and
snapping a faulty coefficient would hide the fault. The supplementary material (Appendix~E)
covers the gap-threshold sensitivity and the re-orthonormalisation that
column-normalising $\Phi$ forces on $\Pi$.

\section{What Canonicity Provides, and What It Does Not}
\label{sec:canonicity}

Reduced Gr\"obner bases provide a canonical representative for an invariant
ideal. Their primary benefit is \emph{decidable equality}: for a fixed monomial
ordering, two invariant sets define the same ideal if and only if their reduced
Gr\"obner bases are identical. This removes ambiguity caused by different
generating sets for the same ideal and permits exact algebraic comparison
between recovered and reference invariant sets. Membership queries are likewise
decided by normal-form reduction rather than by tolerance-dependent residual
thresholds.

We evaluate this property experimentally. Independent recoveries of identical
physical systems reduce to the same canonical form, whereas systems with
changed invariant ideals are distinguished by two-way normal-form reduction.
The resulting decisions agree with the known fault categories: parameter and
observation faults modify the invariant ideal, while integration and step-size
faults preserve it. Residual-based comparisons do not provide this separation,
because they measure deviation from constraints but do not determine whether
the underlying algebraic object has changed.

Canonicalisation alone does not provide minimality, physical correctness, or a
complete diagnostic procedure. A reduced Gr\"obner basis is unique for a fixed
monomial ordering but is not necessarily a minimum-cardinality generating set.
Likewise, equality of invariant ideals certifies only equality of the encoded
constraints; it does not establish that the constraints fully describe the
dynamics or guarantee correct behaviour. The diagnostic capability developed
in Section~\ref{sec:diagnostic} arises from combining canonical invariant
representations with generator-level screening and parameter-level attribution.

Canonicalisation is unnecessary for residual-based downstream uses such as consistency regularisation and shaping (Section~\ref{sec:e2}); it is required only for equality-based diagnosis.

What is exact, and at which stage, is worth stating plainly, since the pipeline mixes exact and approximate steps and the two should not be confused. The reference ideal is exact by construction, given in $\mathbb{Q}$. The invariant of the system under test is not: it is estimated from finite, floating-point trajectories, and recovery returns a numerical direction rather than a rational one (Section~\ref{sec:quotient}). Only once that direction is fitted to a parametric template and the fitted value rationalised to a bounded denominator (supplementary material, Appendix~D) does an exact comparison become possible again. So the equality decision itself, reference ideal against reconstructed ideal, is exact conditional on that reconstruction; it is not a claim that the statistical estimation preceding it is exact.

\section{The Diagnostic}
\label{sec:diagnostic}

The input is a reference generating set $G_{\rm ref}$ for a system whose
invariants are known, together with trajectories from a system under test:
another release of the same simulator, or a simulator suspected of a configuration
error (Fig.~\ref{fig:pipeline}). The procedure is identical across both settings. The supplementary material (Appendix~E) gives both levels
in full.

\subsection{Level 1: screening}
\label{sec:screening}
Each generator's residual is normalised by the natural scale of $p$ on the data.
For a vanishing constraint the normalised residual is
\begin{equation}
\label{eq:screen-vanish}
r_i \;=\; \frac{\mathrm{RMS}\bigl(p_i(x)\bigr)}{\mathrm{RMS}\bigl(\|x\|^{\deg p_i}\bigr)},
\end{equation}
without which a degree-$3$ generator evaluated on velocities of order ten looks
worse than a unit-norm constraint purely through units, and for a conserved
generator, which has no meaningful distance to zero, the numerator becomes the
standard deviation of $p_i$ \emph{along} each trajectory instead:
\begin{equation}
\label{eq:screen-conserve}
r_i \;=\; \frac{\mathrm{std}_{\text{traj}}\bigl(p_i(x)\bigr)}{\mathrm{RMS}\bigl(\|x\|^{\deg p_i}\bigr)}.
\end{equation}
A fixed threshold on that residual is the obvious rule and the wrong one, since
under measurement noise the reference residual rises off machine epsilon to the
noise floor and a fault must then clear a multiple of the \emph{floor} and not
of the \emph{signal}. We test instead against the reference system's
block-to-block variability: both systems' trajectories are split into $32$
contiguous blocks, and for each generator we compute a one-sided Mann--Whitney
$U$ $p$-value for the test residuals being stochastically larger, Bonferroni
corrected across $G_{\rm ref}$; blocks remain contiguous because conservation
drift is defined over trajectory segments rather than shuffled samples. The
resulting \emph{per generator} $p$-values both detect faults and identify the
violated constraint. Blocks cut from one trajectory are not independent
replicates, inflating the level (Section~\ref{sec:limitations}); the same test
over whole trajectories, with an exact permutation null, holds its nominal
level and is reported alongside throughout.

\subsection{Level 2: attribution}
\label{sec:attribution}
Let $p(x;\theta)$ be a template whose coefficients depend symbolically on
physical parameters and $\hat p$ the polynomial recovered from the test system.
Recovery returns a \emph{direction}, so $\hat p$ has an arbitrary scale and,
for a conserved quantity, an arbitrary additive constant as well, since $E$ and
$\alpha E + \beta$ are conserved together. Fitting is carried out modulo that
gauge, by minimising a normalised residual jointly over $\theta$ and $(\alpha,\beta)$:
\begin{equation}
\label{eq:attribution}
\theta^\star \;=\; \arg\min_{\theta,\,\alpha,\,\beta}\;
\frac{\bigl\|\hat{p} - \bigl(\alpha\, p(\cdot;\theta) + \beta\bigr)\bigr\|}
     {\|\hat{p}\|}.
\end{equation} Fitting without the gauge fails because every coefficient of the Acrobot energy contains $m_2$, leaving no parameter-free monomial for normalisation. Real configuration errors
are sparse, so for each parameter in turn we free that one, hold the rest at
specification, and minimise; templates carry more monomials than parameters, so
a wrong hypothesis shows up as a large residual.

The procedure returns one of three verdicts: \textbf{parameter fault}, when one
hypothesis both fits far better than its nearest rival and lands away from its
specified value; \textbf{consistent with specification}, when a generator was
recovered, the best hypothesis fits, and nothing separates it from its rivals,
which healthy systems produce and a naive argmin would misreport as
a fault; and \textbf{not explicable as a parameter fault}, when
recovery finds no decisive spectral gap, the situation a dropped Coriolis term
produces, or when a direction is recovered that no single-parameter hypothesis
fits. Routing that last case to the second verdict is a mistake with a
direction, since a generator no hypothesis explains is evidence against the
specification.

\subsection{Two downstream uses, and why neither needs canonicity}
\label{sec:downstream}
The same generators support two constructions that consume only a residual.
\emph{Consistency regularisation} adds, to the training loss,
\begin{equation}
\label{eq:reg}
\mathcal{L}_{\rm cons} \;=\; \lambda \sum_{i} w_i\,\bigl\|p_i\bigl(f_\theta(s,a)\bigr)\bigr\|^2.
\end{equation} \emph{Potential-based shaping}
cannot use the residual, identically zero in any real environment; the information
is in the \emph{value} of the conserved quantity, so for a target
$E^\star$ we set
\begin{equation}
\label{eq:shaping}
\Phi(s) \;=\; -\bigl|E(s) - E^\star\bigr|,
\end{equation}
and shape in the sense of
\cite{ng1999shaping}. The supplementary material (Appendix~E) gives both in full.

\section{Experiments}
\label{sec:experiments}

Two propositions proved in the supplementary material (Appendix~F) bound what follows:
Proposition~1 gives soundness with the necessity/sufficiency gap,
so the diagnostic is a necessary condition and we conclude only that, while
Proposition~2 gives the population-level target the screen
approximates, a residual reaching the state only through the generator, which
reseeding cannot move, whereas a two-sample test on state distributions has a
statistic that a change of seed moves without moving the physics; it does not
cover the implemented screen, a finite-sample test of a different statistic,
and the healthy controls of Section~\ref{sec:e1} are the evidence there.

Environments come in three tiers (supplementary material, Appendix~G):
\emph{Tier A} has exact vanishing invariants with known ground truth (Acrobot,
whose $6$-dimensional observation is already the coordinate system the algebra
wants, and Reacher, whose $11$ dimensions answer the dimensionality
objection); \emph{Tier B} is actuated and dissipative, where the power
balance replaces conservation; and \emph{Tier C} holds the negative controls,
Hopper and HalfCheetah, on which the low-degree nullspaces are empty at every
tolerance tried and the procedure correctly returns nothing.

The diagnostic is scored against a catalogue of fifteen faults fixed in
advance, never against random perturbations, because a fixed list has a known
\emph{locus}, the generator the fault must break, and for parameter faults a
known \emph{value}, which is what makes localisation and attribution scorable
at all: seven parameter faults (e.g.\ \texttt{acrobot\_m2\_1.3}, mass $1.3$ not
$1.0$), two structural (a dropped Coriolis term), two numerical (the shipped,
unresolved step), and four observation faults (e.g.\ transposed
$\sin\theta_1,\cos\theta_2$), plus two healthy controls on which the correct
behaviour is silence; Table~S6 in the supplementary material (Appendix~G) lists all fifteen with the generator each
must break.

\subsection{Localising and attributing physics faults}
\label{sec:e1}

Both experimental configurations come from an analytic reimplementation of the two
environments, since the shipped ones fix the step size and the integrator while
this experiment has to vary both; it is checked against rollouts of Gymnasium
and the symbolic energy (supplementary material, Appendix~I). The reference
configuration runs at specification, and the test configuration contains one injected fault. Two regimes ask different questions of the
baselines: \emph{unpaired} logs the two systems independently, the situation
anyone comparing two releases is in, and \emph{paired} starts both from
identical initial conditions, the regime the two-sample tests are entitled to.

\begin{table*}[t]
\centering
\captionsetup{font=footnotesize, justification=centering, labelsep=period}
\caption{Detection, localisation and attribution over fifteen faults and two
healthy controls, one draw per cell. Localisation is scored as exact set
agreement with the fault's known constraint locus, and \emph{n/a} marks a method
with no constraint to name or, for the paired difference test, a regime it
cannot run in. The screen appears at two units of replication, blocks cut from
each trajectory and whole trajectories under an exact permutation null. The
paired noiseless control column is an identity check rather than a false-alarm
rate; Fig.~\ref{fig:calibration} measures the rate.}
\label{tab:e1}
\renewcommand{\arraystretch}{1.15}
\small
\begin{tabularx}{\textwidth}{|l|l|*{4}{>{\centering\arraybackslash}X}|}
\hline
\textbf{Regime, noise} & \textbf{Method} & \textbf{Faults detected} & \textbf{False alarms} & \textbf{Localised} & \textbf{Attributed} \\
\hline
\multirow{6}{*}{\parbox{1.6cm}{\centering\scriptsize unpaired,\\$\sigma\!=\!0$}}
  & MMD permutation      & $12/15$ & $1/2$ & n/a & n/a \\
  & KS $+$ Bonferroni    & $10/15$ & $1/2$ & n/a & n/a \\
  & MLP discriminator    & $12/15$ & $1/2$ & n/a & n/a \\
  & paired difference    & n/a     & n/a   & n/a & n/a \\
  & \cellcolor{blue!12}\textbf{algebraic, per block} & \cellcolor{blue!12}$\mathbf{15/15}$ & \cellcolor{blue!12}$\mathbf{0/2}$ & \cellcolor{blue!12}$\mathbf{15/15}$ & \cellcolor{blue!12}$7/7$ \\
  & \cellcolor{blue!12}\textbf{algebraic, per trajectory} & \cellcolor{blue!12}$\mathbf{15/15}$ & \cellcolor{blue!12}$\mathbf{0/2}$ & \cellcolor{blue!12}$\mathbf{15/15}$ & \cellcolor{blue!12}$7/7$ \\
\hline
\multirow{6}{*}{\parbox{1.6cm}{\centering\scriptsize paired,\\$\sigma\!=\!0$}}
  & MMD permutation      & $6/15$  & $0/2$ & n/a & n/a \\
  & KS $+$ Bonferroni    & $6/15$  & $0/2$ & n/a & n/a \\
  & MLP discriminator    & $8/15$  & $0/2$ & n/a & n/a \\
  & \cellcolor{green!12}\textbf{paired difference} & \cellcolor{green!12}$\mathbf{15/15}$ & \cellcolor{green!12}$0/2$ & \cellcolor{green!12}n/a & \cellcolor{green!12}n/a \\
  & \cellcolor{blue!12}\textbf{algebraic, per block} & \cellcolor{blue!12}$\mathbf{15/15}$ & \cellcolor{blue!12}$0/2$ & \cellcolor{blue!12}$\mathbf{15/15}$ & \cellcolor{blue!12}$7/7$ \\
  & \cellcolor{blue!12}\textbf{algebraic, per trajectory} & \cellcolor{blue!12}$\mathbf{15/15}$ & \cellcolor{blue!12}$0/2$ & \cellcolor{blue!12}$\mathbf{15/15}$ & \cellcolor{blue!12}$7/7$ \\
\hline
\multirow{6}{*}{\parbox{1.6cm}{\centering\scriptsize paired,\\$\sigma\!=\!10^{-3}$}}
  & MMD permutation      & $6/15$  & $0/2$ & n/a & n/a \\
  & KS $+$ Bonferroni    & $7/15$  & $0/2$ & n/a & n/a \\
  & MLP discriminator    & $8/15$  & $0/2$ & n/a & n/a \\
  & \cellcolor{green!12}\textbf{paired difference} & \cellcolor{green!12}$\mathbf{15/15}$ & \cellcolor{green!12}$\mathbf{0/2}$ & \cellcolor{green!12}n/a & \cellcolor{green!12}n/a \\
  & algebraic, per block & $11/15$ & $0/2$ & $10/15$ & $2/7$ \\
  & algebraic, per trajectory & $12/15$ & $\mathbf{0/2}$ & $\mathbf{12/15}$ & $2/7$ \\
\hline
\end{tabularx}
\end{table*}

The screen detects $15/15$ and localises $15/15$ at $\sigma=0$ in both the
unpaired regime and the paired identity check,
with no observed false alarm on either healthy control. It does not attribute the detection:
a paired two-sample test built from the same information, per-coordinate
block-mean discrepancy against a healthy paired pair, also detects $15/15$ at
both noise levels and at $\sigma = 10^{-3}$ detects what the screen misses.

The localisation column is where the two differ. No baseline names a
constraint: MMD \cite{gretton2012mmd} and the discriminator, a classifier
two-sample test in the sense of \cite{lopezpaz2017c2st}, return one number for
the whole state vector, while KS and the paired difference test could name an offending
\emph{coordinate}, which is not a constraint, and on a parameter fault every
coordinate moves at once. In the unpaired regime all three of them fire on the
reseeded Acrobot control, a different seed being a different state distribution
that a distributional test cannot separate from different physics, and the
screen fires on neither control. At $\sigma = 10^{-3}$ the screen detects
$12/15$ per trajectory and still localises what it detects.

That comparison gives the screen named features and the baselines none, which
confounds two differences. Handed the per-generator residual vector instead of
the raw observation, the baselines localise too, an off-the-shelf
Kolmogorov--Smirnov test over it reaching $14/15$. Localisation is therefore
provided by the reference generating set and not by the rank test reading it, and
what exactness and canonicity supply is the set itself, well defined,
reproducible across runs and comparable between two systems.

Counts at one fault magnitude also say nothing about where a method's floor
lies. Sweeping one constant at a time from $10^{-4}$ to $0.3$ places the
screen's noiseless floor at the bottom of the sweep on all four constants, where
no baseline resolves better than $30\%$ on Acrobot; under noise it holds
$10^{-3}$ on Reacher and rises by two to three orders on Acrobot, whose energy
is a drift measured over a window that observation noise enters directly. Both
ablations and the operating characteristics over $\alpha$ are in
the supplementary material (Appendix~H), and
Fig.~\ref{fig:calibration} measures the false-alarm rate over $500$ pairs per
setting.

Attribution names the correct parameter on all $7$ parameter faults, to a
relative error of $1.0\times10^{-7}$, returns \emph{not explicable as a
parameter fault} on all seven faults that have a parameterised generator and
that no parameter explains, and returns \emph{consistent with specification} on
both healthy controls at $\sigma = 0$. It reads coefficients off a singular
direction, so its accuracy depends on that direction's conditioning, and the conserved-quantity
half fails between $\sigma = 10^{-6}$ and $10^{-5}$
(Table~S8 in the supplementary material). The consequence is a division of labour, stated
here as a limitation: level-2 attribution serves simulators and version diffs,
where data is float64 and noise-free, and not logged hardware, whereas level-1
screening needs only a residual and stays useful at noise three orders of
magnitude higher.

\subsection{Does exact consistency improve a world model?}
\label{sec:e2}

We train eighty-four MLP dynamics models on Acrobot, varying $\lambda$, hidden width, trajectory budget and seed. The sweep extends to $\lambda=100$, where one-step validation error increases $64$-fold while the algebraic residual falls to $0.79$ of its unregularised value, ensuring the regulariser is active.

Over that whole range no weight improves the divergence horizon
(Table~S11 in the supplementary material). Every point estimate is at or below zero, the largest
of them $-0.1$ steps out of $45.9$, and the one bootstrap interval that excludes
zero does so on the losing side, $\lambda = 10^{-2}$ at $-1.3$ steps. Past
$\lambda = 1$ the horizon collapses at every threshold from $0.05$ to $1.0$,
losing $37$ steps at $\lambda = 100$.

One-step validation error is the better predictor of long-horizon fidelity, and
the pool it is measured over changes how much better. On the unregularised
models Spearman is $0.93$ for validation error against $0.77$ for the algebraic
residual; the ordering survives on every pool and the margin does not
(supplementary material, Appendix~H). Validation error also needs the held-out
ground truth the algebraic residual does not. The exact invariant does not
outperform the approximate construction it was meant to improve on, because the
loss consumes only a residual, and a residual does not care whether its
generator is canonical.

\begin{figure*}[t]
\begin{center}
\includegraphics[width=\textwidth]{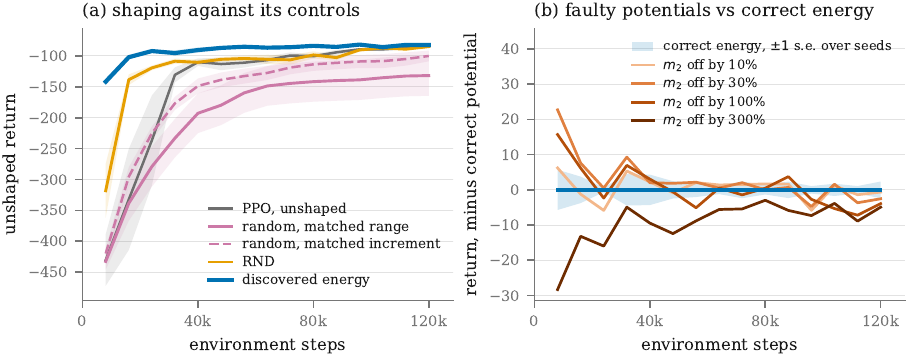}
\end{center}
\caption{Acrobot swing-up, unshaped environment return, mean over five seeds.
\emph{(a)} The discovered energy leads unshaped PPO for the whole of the first
$40$k steps. The two random configurations, of the same degree and matched in
scale two different ways, each contain ten independent polynomial draws and five
PPO seeds per draw; all remain behind the discovered potential.
\emph{(b)} Each faulty potential is plotted against the correct one so that the
comparison is against seed-to-seed spread and not against the axis. Mass errors
of $10\%$, $30\%$ and $100\%$ stay inside the band; only $300\%$ leaves it, and
it leaves it downwards for most of training.}
\label{fig:shaping}
\end{figure*}

\subsection{Shaping, and the control that decides what it means}
\label{sec:e3}

Shaping from the discovered energy is a large sample-efficiency gain, improving
AUC from $-147.8$ to $-91.3$, and among the energy-derived conditions policy
invariance holds empirically as \cite{ng1999shaping} predicts, each converging
to a final return within about one standard deviation of the others; the gain is
in how fast the policy arrives. A random-polynomial control, matched in degree
and scale to the discovered potential under two normalisations
(supplementary material, Appendix~H), stays behind it for the whole run, so the
gain is not simply from adding a dense shaping signal.

The faulty configurations gauge themselves entirely inside the faulty model,
with nothing correct entering at any step, which is the position of a
practitioner deploying from a model they have not checked. They still shape.
At $10\%$, $30\%$ and $100\%$ mass error the AUC is $-90.8$, $-88.6$ and
$-90.8$ against the correct potential's $-91.3$, all three inside its own
seed-to-seed spread for most of training (Fig.~\ref{fig:shaping}b) and each
difference under a tenth of the $-56.5$ that shaping gains over none
(bootstrap intervals in supplementary Appendix~H). Only at $300\%$ does
the potential degrade, to $-100.5$, and even there it beats no shaping.

That robustness argues for the diagnostic and not against it. Learning absorbs a
parameter error without reporting one, so a training run is a poor instrument
for detecting the error it tolerates, and the cost lands elsewhere: on a policy
that transfers to hardware built for the right mass, and on a return quoted
against a benchmark whose physics has moved. Section~\ref{sec:e4} finds two such
defects in shipped code, neither of which changes a reported return.

\subsection{Auditing shipped simulator releases}
\label{sec:e4}

Every fault above was injected by us, the standard objection to a
mutation-based evaluation, so we answer it by pointing the same pipeline,
unchanged, at the release history of \texttt{gym}, \texttt{gymnasium} and the
\texttt{mujoco} bindings: $350$ pairs over eleven environments, each pinned
into its own virtualenv and built through the public \texttt{make} entry
point with no source file patched. A dumper running inside each virtualenv
imposes identical initial states and action sequences, so unchanged dynamics
code on byte-identical inputs must reproduce byte-identical doubles; the
exact stage decides detection outright, and the screen runs on the pairs it
separates to decide localisation.

The dynamics are unchanged throughout. All $275$ classic-control pairs are
bitwise identical, and among the $75$ MuJoCo pairs the four non-chaotic
environments agree to at most $9.1\times10^{-15}$, the signature of a recompiled
physics library reordering floating-point operations. The null is the expected
answer on maintained projects. What this experiment contributes
is not the null but three properties of the benchmarks that no source comparison
would have surfaced, none of them being a line anyone changed. The pipeline is
not blind while returning them: perturbing Gymnasium's \texttt{AcrobotEnv},
the exact stage separates both controls on byte-identical inputs
(supplementary material, Appendix~H). Whether the screen localises them at the
shipped step size is a separate question; the supplementary material reports where
localisation is demonstrated instead.

\paragraph{Reacher's observation fails its own forward kinematics.} On Reacher
the forward-kinematics generators sit at $1.1\times10^{-8}$ and
$1.6\times10^{-7}$ in a float64 pipeline, nine orders above the precision the
unit-norm identities reach in the same observation vector, and the geometry is
not the cause. \texttt{MujocoEnv.\_step\_mujoco\_simulation} calls
\texttt{mj\_step} and never \texttt{mj\_forward} afterwards, so
\texttt{data.xipos} still describes an internal integrator stage while
\texttt{ReacherEnv.\_get\_obs} takes $\cos\theta$ and $\sin\theta$ from the
updated \texttt{qpos} and the fingertip vector from the stale \texttt{xipos}:
the two halves of the observation are read at different points of the step. At
$6\times10^{-9}$ this changes nobody's returns and it is not a physics error,
but it is a real inconsistency in a widely used benchmark, it floors any exact
diagnostic on shipped MuJoCo observations at about $10^{-8}$, and it reproduces
on all six installed configurations and both Reacher versions. Inserting a single
\texttt{mj\_forward} call after \texttt{mj\_step} in the patched simulation loop
drops the FK residual from $6.2\times10^{-9}$ to $2.8\times10^{-17}$, matching
the geometry baseline and confirming the diagnosis. What surfaced it
was the per-generator report: two generators of one reference set, on one
observation vector, differing by nine orders of magnitude, whereas an aggregate
score returns one unremarkable number.

\paragraph{Acrobot ships at a step size where energy is not usable.} A one
percent error in \texttt{LINK\_LENGTH\_1} moves the trajectory by
$1.6\,\mathrm{rad}\,\mathrm{s}^{-1}$, which the exact stage sees immediately
and the screen does not flag: Gymnasium ships Acrobot at $\Delta t = 0.2$,
where RK4's relative energy drift of $3.55\times10^{-3}$ is a larger violation
of conservation than a percent-level parameter error. Reducing the step at
fixed horizon (Table~\ref{tab:auditdt}) drops the drift floor and the smallest
localisable parameter error together, from nothing localised at the shipped
step to below $10^{-3}$ at $\Delta t = 0.05$: energy is not a usable invariant
of Acrobot as distributed, and any diagnostic relying on it has that same
floor.

\begin{figure}[t]
\begin{center}
\includegraphics[width=\columnwidth]{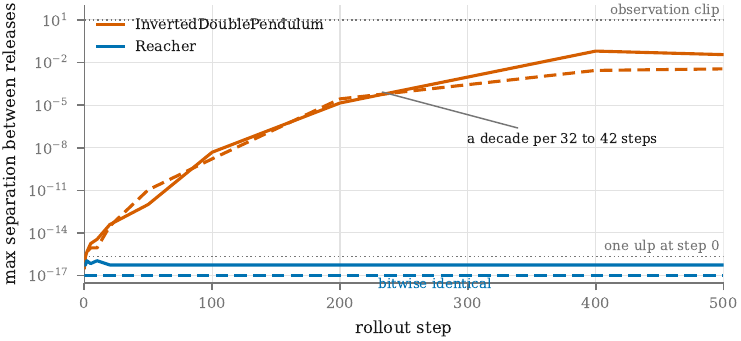}
\end{center}
\caption{Separation between two shipped releases running identical model files,
from a state both were handed identically. Colour is the environment and line
style the release pair. The chaotic environment climbs fourteen decades at a
constant rate from one unit in the last place, which is the signature of an
amplified rounding difference and not of a changed model; the non-chaotic one,
over the same release pairs and the same horizon, does not move. The lower
Reacher pair is bitwise identical at every step and is drawn on the axis
floor.}
\label{fig:amplification}
\end{figure}

\paragraph{On a chaotic benchmark an exact comparison must be read at the first
step.} On InvertedDoublePendulum, the one chaotic environment in the matrix,
thirteen release pairs separate by between $4.5\times10^{-2}$ and the $10^{1}$
at which the observation saturates its clip; read at that horizon it would
look like the largest dynamics change in the audit, and it is not one. The
same pairs differ by at most $2.2\times10^{-16}$ on the first step, from a
state both releases were handed identically, and the separation grows at a
decade per $32$ steps (Fig.~\ref{fig:amplification}), the system's Lyapunov
exponent and not a changed model; Reacher over the same pairs and horizon does
not separate at all. The exact stage's guarantee survives only where read
before the dynamics amplify anything, so a pair is classified by its
first-step separation, and an audit thresholding the horizon would report a
physics change in a benchmark whose physics did not change.

\begin{figure}[t]
\begin{center}
\includegraphics[width=\columnwidth]{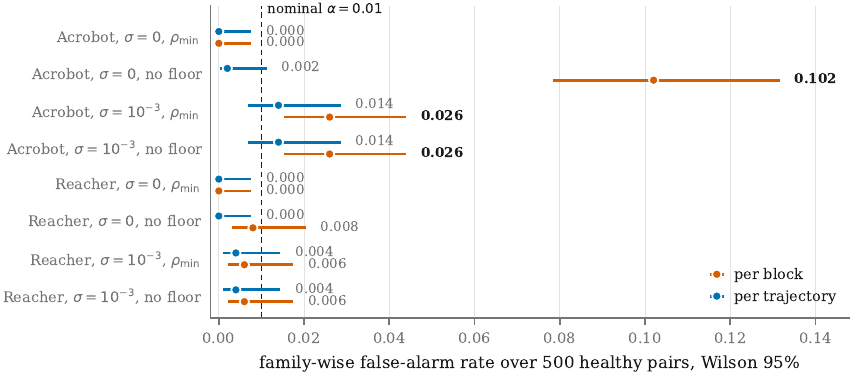}
\end{center}
\caption{Measured family-wise false-alarm rate over $500$ healthy-versus-healthy
comparisons per setting, against a nominal $\alpha = 0.01$, with Wilson $95\%$
intervals. Both configurations are the same model differing only by seed, so every
rejection is a false alarm by construction. Three intervals lie
entirely to the right of the nominal line and all three are block-level and on
Acrobot; every trajectory-level interval covers it or sits below.}
\label{fig:calibration}
\end{figure}

\begin{table}[t]
\caption{Release audit. The reference system's energy drift sets the floor,
and the
smallest relative parameter error the screen localises tracks it. The first row
is the shipped configuration. Both parameters have spec value $1.0$, and
$\le 10^{-3}$ means every magnitude down to the bottom of the sweep was
localised, and ``none'' means no swept magnitude was. At the shipped step size
the energy's own drift floor is a larger violation of conservation than a
$30\%$ parameter error, which is Remark~2 in the supplementary material stated as a
measurement.}
\label{tab:auditdt}
\begin{center}
\footnotesize
\setlength{\tabcolsep}{2pt}
\begin{tabular}{llll}
$\Delta t$ & reference energy drift & \texttt{LINK\_MASS\_2} &
  \texttt{LINK\_LENGTH\_1} \\
\hline
$0.2$ (shipped) & $3.55\times10^{-3}$ & none & none \\
$0.05$          & $4.19\times10^{-6}$ & $\le 10^{-3}$ & $\le 10^{-3}$ \\
$0.01$          & $1.52\times10^{-8}$ & $\le 10^{-3}$ & $\le 10^{-3}$ \\
\end{tabular}
\end{center}
\end{table}

\section{Limitations}
\label{sec:limitations}

The method has several scope conditions. Attribution requires sufficiently clean trajectories and, in the present implementation, rational reference parameters, bounded-degree polynomial invariants, and sparse monomial support. Its intended setting is therefore controlled simulator and version-difference analysis rather than arbitrary logged hardware. Multiple simultaneous faults are also outside the single-parameter attribution setting used here; freeing multiple parameters can recover more than one fault, but this is not the default search problem.

The screening stage requires a reference generating set, so the method is not reference-free; this is most relevant for first releases, hand-derived models, or other settings in which a trusted specification exists but the implementation may not. When the correct parametric template is already known and only its constants are in question, directly fitting that template is a simpler route to the same estimate, and we do not claim an accuracy advantage over it in that case. What a direct fit cannot do is decide whether a single-parameter explanation exists at all: fitting a template presupposes the template applies, so nothing in the fit itself can signal that it does not. Attribution can, and already does so in Section~\ref{sec:e1}, returning \emph{not explicable as a parameter fault} on every one of the seven faults with a parameterised generator that no single constant explains, and \emph{consistent with specification} on both healthy controls, rather than forcing a best-fit constant onto data a template does not describe. That refusal, not accuracy on the parameter faults, is what discovery is for.

The numerical stages are also constrained by simulator fidelity. Discovery is affected by integrator drift, while long-horizon exact comparison is limited by dynamical amplification in chaotic systems; contact-rich locomotion is outside the present construction. These effects mean that sufficiently fine re-integration or short-horizon comparison may be necessary when the shipped simulator itself is numerically limiting. Finally, the block-level rank test does not attain its nominal false-alarm level when blocks from a single trajectory are treated as independent replicates, so the trajectory-level permutation test is the unit on which the stated error control is based.

\section{Conclusion}
\label{sec:conclusion}

We tested a premise assumed throughout the physics-informed learning
literature, that enforcing physical structure improves learned dynamics. For
prediction it did not hold: increasing the regularisation weight until it
cost the model a factor of sixty in one-step error produced no improvement in
rollout fidelity, and a shaping potential recovered and gauged entirely
within a system with a $100\%$ mass error performed as well as the correct
one. Canonicity provides decidable equality and reproducibility, not
minimality, sufficiency, or safety, so the applications worth building are
diagnostic: given trajectories from a simulator or release, it reports which
constraint broke and, if the fault is a wrong constant, what it has become.
Turned on the field's infrastructure, it finds no change across $350$
release pairs, and three benchmark properties no comparison would find.

\section*{Reproducibility Statement}

Code for every experiment is available in \url{https://www.github.com/TesfayZ/algebraicRLtest.git}. Every number in Section~\ref{sec:experiments}
and supplementary Appendix~I is produced by a script there; the mapping from result to script is in the accompanying README,
and the checks of supplementary Appendix~I exit nonzero if any claim there
has drifted from the text.

\bibliographystyle{IEEEtran}
\bibliography{references}

\end{document}


\maketitle

\noindent This document contains the appendices for ``Diagnosing Faults in Reinforcement Learning Simulators with Canonical Polynomial Invariants.'' References such as Section~V or Table~I of the main paper point to the main paper; tables, figures and equations in this document carry the prefix S, and appendices are lettered A--I.

\appendix

\section{Extended Related Work}
\label{app:related}

Symbolic regression (e.g., \cite{cranmer2023pysr, udrescu2020aifeynman}) and SINDy (e.g., \cite{brunton2016sindy}) are well established; our front end uses a sparse support search. What follows is the part of the literature the diagnostic's
claims rest on.

\paragraph{Discovery without a canonical form.} PySR \citep{cranmer2023pysr},
AI Feynman \citep{udrescu2020aifeynman}, SINDy \citep{brunton2016sindy} with its
weak-form \citep{messenger2021weaksindy_ode} and implicit
\citep{kaheman2020sindypi} variants, and the conservation-law methods of
\citet{kaiser2018discovering} and \citet{oellerich2024robust} all return
floating-point objects with no canonical form, so none supports equality between
two discovered invariant sets. We utilized AI Feynman's nondimensionalisation in
Section~\ref{sec:units}, where it is necessary and not merely convenient.

\paragraph{Border bases are not the same currency.} AVI
\citep{heldt2009approximate} and Vanishing Component Analysis
\citep{livni2013vca} compute border bases of vanishing ideals from noisy
samples; \citet{kera2020gradient} remove spuriously vanishing and symbolically
redundant elements by a gradient criterion, addressing the redundancy our
normal-form reduction removes but numerically and not by ideal membership, and
\citet{wirth2023avi} scale oracle-based AVI to large samples. The
difference from our setting is fundamental: a border basis is not simply a
less convenient canonical form. Membership in an approximate border basis is a
thresholding decision, whether a residual is small enough. Membership in an
ideal presented by a reduced Gr\"obner basis is a decidable predicate,
$\NF_G(q) = 0$, evaluated in exact rational arithmetic
\citep[Chapter 2, Section 6]{cox2015ideals}. Every claim the diagnostic makes
about two systems having or not having the same algebra depends on the second
and cannot be made with the first at any tolerance; this is the point argued in
Section~V of the main paper. The cost is that the coefficients must be
rational, which is why the reference side reduces to an exact form and the test side does not
(Section~IV-B of the main paper). Numerical algebraic geometry represents a variety
by witness sets from homotopy continuation instead of by generators
\citep{sommese2005numerical}, answering geometric questions we do not ask and
delivering no syntactic equality test between two generating sets;
\citet{breiding2018learning} study learning varieties from samples.

\paragraph{Invariants inferred from executions.} Dynamic invariant detection
\citep{ernst2007daikon} is the closest precedent to the screen: it runs a
program, watches the values its variables take and reports the relations that
held throughout, and it has been used to compare one version of a program
against another. Three differences decide why it cannot do what
Section~VII-D of the main paper does. Its candidate relations come from a fixed template
library over program variables, so the hypothesis class is enumerated in advance
where ours is a degree-bounded polynomial space explored by sparse support
search. Its output is
annotated with a confidence that the relation is not accidental, a statistical
statement about coverage, where a reduced Gr\"obner basis over $\mathbb{Q}$ is
an exact object. And two of its inferred sets support no equality test, so
comparing versions there is a diff of two lists of strings, which is sensitive
to how each list happened to be written; the equality decision of
Section~V of the main paper is invariant to that. Property differencing
\citep{yang2014propertydiff} takes the last point seriously and makes the
difference between two versions' expected properties the object checked, but the
properties are supplied by the developer rather than recovered from behaviour,
which is the case the audit cannot assume: nobody wrote down what Acrobot's
energy should be before we did.

\paragraph{Deflation.} \citet{zhu2023neuraldeflation} discover functionally
independent conservation laws by deflating a neural loss against
already-recovered invariants. Our normal-form reduction procedure is the exact
algebraic analogue: reducing modulo a reduced Gr\"obner basis removes the trivial
multiples exactly and certifiably, instead of penalising them approximately,
and the same operation yields the canonical form. A common algebraic deflation
approach uses orthogonal projection of the nullspace basis, which removes only
the span of the known generator. Our normal-form reduction, by contrast,
removes the entire filtered component of the ideal, which is the structural
distinction that makes the deflation effective in the diagnostic setting.

\paragraph{Enforcing invariants in learned dynamics.} ConCerNet
\citep{zhang2023concernet} is the closest prior work to
Section~VI-C of the main paper, discovering invariants contrastively and enforcing
them with a neural projection layer. The difference is what the invariant
\emph{is}. ConCerNet's is a learned latent function, so it cannot be inspected,
cannot be handed to a solver, and admits no notion of equality between two
discovered sets. Ours is an exact rational polynomial with a canonical form,
which is what makes the diagnostic possible at all. Structured architectures,
Hamiltonian \citep{greydanus2019hamiltonian} and Lagrangian networks
\citep{cranmer2020lagrangian,lutter2019delan} among them, assume the conserved
structure instead of discovering it, and assume a form that actuated dissipative
systems do not have. SINDy-RL \citep{zolman2025sindyrl} learns sparse symbolic
dynamics explicitly, which presupposes the dynamics are sparsely representable,
where the diagnostic treats $f_\theta$ as a black box and asks only whether its
trajectories satisfy the algebra; a sparse symbolic model is one more artefact
it could be pointed at.

\paragraph{World models, and what the world-model experiment evaluates.} Learned
dynamics models are trained and then rolled out under their own predictions,
from the recurrent world models of \citet{ha2018worldmodels} to the family
culminating in \citet{hafner2025dreamer}, and what decides whether such a model
is usable is long-horizon rollout fidelity and not one-step error. That is
exactly the setting in which no held-out ground truth is available, so a
reference-free score is useful even when it is not the best predictor,
which is the finding of Section~VII-B of the main paper. Our world model is a
deliberately plain MLP, because the question is whether an exact algebraic
penalty changes rollout fidelity at all, and a latent-space model would confound
that with its representation's error.

\paragraph{Simulator fidelity, and testing it.} The dominant response to a
simulator that does not match its target is to randomise over the mismatch,
visually \citep{tobin2017domainrand} or over the dynamics parameters themselves
\citep{peng2018simtoreal}. The two objectives compose: randomisation widens the
distribution a policy tolerates, whereas the screen decides whether a specific
parameter is wrong and attribution says what it is. Randomisation also
presupposes that the parameters being randomised are the ones that matter and
that the equations around them are right, the assumption a structural fault
violates and the screen tests directly. Metamorphic testing
\citep{chen2018metamorphic} checks software by relations that must hold between
inputs and outputs instead of by known correct answers, which is structurally
what we do, except that its relations are hand-written and ours are discovered.
Property-based testing of numerical code and the literature on validating
scientific simulators \citep{oberkampf2010verification} share the goal, and
reproducibility studies \citep{henderson2018deeprl} document how often results
move for reasons nobody can localise. Version drift between releases is a
recognised instance, and we are not aware of prior work returning a symbolic
account of what changed.

\paragraph{System identification, and why this is not that.} The obvious
objection to attribution is that estimating a mass from trajectories is
classical system identification, fitting the parameters of a known dynamics
model by least squares \citep{ljung1999sysid}. The objection is fair, and it identifies an important boundary of the method. Direct system identification is the simpler route when the correct parametric dynamics model is already known and only its parameters are unknown. Our setting is broader: the relevant invariant need not be supplied in advance, and attribution can decline when the recovered invariant does not admit the assumed parameter explanation. A structural fault such as a dropped Coriolis term can therefore fall outside the assumed template rather than being forced into a best-fit parameter. Our level-2 fit is against the parametric form of an invariant, a weaker model commitment, and it reports non-identifiability explicitly (Section~VI-B of the main paper). Level-1
screening, for its part, presupposes no parametric model at all. And identification needs logged actions and a simulator in the loop, whereas screening needs only states.

\paragraph{Shaping, safety and Gr\"obner bases.} Our shaping term is
potential-based, so \citet{ng1999shaping} applies directly and leaves the
optimal policy unchanged; Section~VI-C of the main paper explains why the more
obvious construction, an intrinsic reward penalising the invariant residual, is
identically zero in any real environment. We compare against RND
\citep{burda2019rnd} and two controls of our own. For safety, barrier
certificates \citep{prajna2004barrier} and shielding
\citep{alshiekh2018shielding} are the standard routes. Gr\"obner bases
have an established role in \emph{invariant generation for formal
verification}: \citet{sankaranarayanan2004groebner} generate non-linear
polynomial invariants of transition systems, \citet{rodriguezcarbonell2007}
characterise all polynomial invariants of simple loops, and DIG
\citep{nguyen2014dig} generates them dynamically from execution traces. All
three are given the transition relation symbolically, whereas ours is a learned
network, so the invariants must come from data and the verification query must
account for model error. Section~V of the main paper argues that invariants can
shrink that error set but cannot separate safe from unsafe, and we therefore do
not pursue shielding.

\section{Vanishing Ideals, Conserved Algebras and Balance Laws}
\label{app:prelim}

This section fixes a distinction that is easy to elide, and that has
consequences for every subsequent claim once elided.

\subsection{Vanishing constraints form an ideal}
\label{sec:ideal}

Let $\mathcal{M} \subseteq \R^n$ be the manifold of physically admissible
states. The set
$$
I(\mathcal M) \;=\; \{\, p \in \R[x_1,\dots,x_n] \;:\; p(x) = 0
  \ \ \forall x \in \mathcal M \,\}
$$
is an ideal: it is closed under addition and under multiplication by arbitrary
polynomials. Section~III of the main paper lists the examples; two have detail worth
recording here. Rigid-link constraints take the form
$(x_2-x_1)^2 + (y_2-y_1)^2 - L^2$, and stoichiometric moiety conservation
$\sum_i \nu_i c_i - N$ in reaction networks
\citep{schuster1991conservation} is linear with integer coefficients, which puts
it within reach at any dimension.

\begin{remark}[The reduced Gr\"obner basis is not a minimal generating set]
\label{rem:notminimal}
Uniqueness is often mistaken for minimality. It is not. For
$I = \langle y - x^2,\ z - x^3 \rangle$ (the twisted cubic), the reduced
Gr\"obner basis under grevlex is $\{\,x^2 - y,\ xy - z,\ y^2 - xz\,\}$: three
elements for an ideal with two generators, the third being an algebraic
consequence of the first two (verified symbolically,
Appendix~\ref{app:tc}). Consequently a loss of the form
$\sum_{p \in G} \|p(\cdot)\|^2$ does \emph{not} assign one penalty per
independent constraint. Separately, even a minimal generating set need not
have one generator per unit of codimension, since most varieties are not
complete intersections. Where minimality is genuinely required we obtain it
by an explicit extra step (interreduction, or minimal generators via the
degree-graded strand of a minimal free resolution for homogeneous ideals) and
we say so; we do not obtain it from canonicalisation.
\end{remark}

\subsection{Conserved quantities form an algebra, not an ideal}
\label{sec:algebra}

A conserved quantity of the dynamics $\Phi$ satisfies $p(\Phi(s)) = p(s)$. The
set of all such $p$ is the fixed ring of the endomorphism
$p \mapsto p \circ \Phi$, closed under addition and, since $E$ conserved implies
$E^2$ conserved, under multiplication as well, hence a \emph{subalgebra} of
$\R[x]$ and not an ideal. (The non-trivial conserved quantities, those with
$p \not\equiv 0$, do not themselves form a subalgebra, $0$ being excluded; the
object with the algebraic structure is the full fixed ring.) Reduced Gr\"obner
bases canonicalise ideals, so they do not apply to it as stated, and
Section~III of the main paper gives the practical failure that follows. Two changes
address it.

\paragraph{Difference dictionary.} Search over
$\mathcal{D}_{\mathrm{diff}} = \{\, m(s') - m(s) \;:\; m \text{ a monomial with
} 1 \le \deg m \le d \,\}$, whose cardinality is that of the degree-$d$
monomial dictionary in $n$ variables rather than in $2n$. The constant monomial
is excluded deliberately: its difference column is identically zero, so leaving
it in would hand every system a spurious exact nullspace vector and make the
refusal of Section~IV-B of the main paper unreachable. A vector in the nullspace of
the
associated design matrix is a polynomial $p$ with $p(s') = p(s)$ on the data,
i.e.\ a conserved quantity, and it is recovered at every energy level
simultaneously.

\paragraph{Functional independence in place of ideal minimality.} Extract a
maximal functionally independent subset, certified by the rank of the Jacobian
$J = [\nabla p_1; \dots; \nabla p_k]$ at sampled regular points. This discards
$E^2$ given $E$, since $\nabla(E^2) = 2E\nabla E$ is parallel to $\nabla E$,
and that is the redundancy the ideal machinery handles in the vanishing case.

A maximal functionally independent subset is not a canonical form, and the
difference matters here. It is not unique, since
$\{E_1, E_2\}$, $\{E_1, E_1 + E_2\}$ and $\{E_1, E_1E_2\}$ are all maximal
independent sets for the same algebra; it depends on the order in which
candidates are examined and on the points sampled; and it is not invariant under
a change of generators. What it delivers is an irredundancy certificate. The
canonical-form question for subalgebras is answered by SAGBI bases
\citep{robbiano1990sagbi,kapur1989sagbi}, which are the subalgebra analogue of a
Gr\"obner basis, but a finitely generated subalgebra need not admit a finite
SAGBI basis under any given order, so there is no counterpart to the
finiteness that makes the ideal side decidable. We therefore claim canonicity
for the vanishing ideals only, and deduplication for the conserved
quantities. The same rank computation reappears in
Section~\ref{sec:sufficiency} as the sufficiency diagnostic; it does double
duty.

\subsection{Balance laws: the actuated case}
\label{sec:balance}

For an actuated, damped Lagrangian system, energy is not conserved, but the
power balance
$$
\frac{dE}{dt} \;=\; u^\top \dot q \;-\; \dot q^\top D \dot q
$$
is an exact identity, and both sides are polynomial in $(\dot q, u)$ for
constant damping $D$. Discretised, this is a polynomial relation on the
transition graph in the joint space $(s, a, s')$:
\begin{equation}
\label{eq:balance}
E(s') - E(s) - \Delta t\,\big(u^\top \dot q - \dot q^\top D \dot q\big)
  \;=\; O(\Delta t^{2}),
\end{equation}
evaluating the power at $s$. Which quadrature is used decides whether the
relation is recoverable at all, so it is not a presentational choice. The
rectangle form above is the obvious discretisation and the wrong one to search,
its defect being first order in $\Delta t$ and landing in the same nullspace the
damping coefficient occupies; averaging the power at the two endpoints gives a
defect of $O(\Delta t^3)$, and that trapezoid form is what the experiments use.
Appendix~\ref{app:environments} measures the difference. Either is
recoverable with the dictionary
$\{m(s') - m(s)\} \cup \{\Delta t \cdot m(s,a)\}$, of size
$\binom{n+d}{d} + \binom{n+m+d}{d}$ against the $(2n{+}m)$-variable
alternative, which is roughly a doubling when the action dimension $m$ is small
($204$ against $84$ for Acrobot at $d=3$). Note what the defect term means:
unlike the vanishing constraints, this relation is exact only in
continuous time, and on the transition graph it holds only to the order of the
quadrature. It
therefore carries a drift floor of its own, and recovering it by a nullspace
method needs a tolerance above that floor, which is the one place
Section~IV-B of the main paper's argument does not reach. This is
how we obtain a usable algebraic relation for a system with no conservation
law, and it is what Section~VI-C of the main paper constrains a world model of a
controlled system to satisfy. We verify numerically on Acrobot that the
residual of \eqref{eq:balance} scales as $\Delta t^2$
(Appendix~\ref{app:balance}), confirming a first-order identity in the joint
space.

\section{Scaling the Search: Deflation, Lifting and Conditioning}
\label{app:scaling}

The diagnostic of Section~VI of the main paper presupposes that the reference
generators can be obtained in the first place, and a criticism of the underlying
discovery method is that it is limited to roughly eight variables while robot
state spaces are larger. This section is what allows the rest of the paper to
apply
at Reacher's eleven dimensions and not only at Acrobot's six. The limit is
real but it is not a limit on $n$, and it is not a Gr\"obner limit. Three things
scale separately.

\paragraph{Dictionary size.} The number of monomials of degree $\le d$ in $n$
variables is $M = \binom{n+d}{d}$: $78$ for $(n,d)=(11,2)$, $364$ for
$(11,3)$, $1365$ for $(11,4)$ (Table~\ref{tab:dict}). Forming and
decomposing an $N \times M$ design matrix at these sizes is inexpensive.

\paragraph{Support search.} Locating a $k$-sparse support among $M$ candidates
is combinatorial, $\binom{M}{k}$ before pruning. \emph{This} is the bottleneck,
and it is driven by $M$, hence jointly by $(n,d)$; degree matters at least as
much as dimension.

\paragraph{Conditioning and sample complexity.} Reliable nullspace estimation
under noise needs $N \gg M$, with the recoverable spectral gap shrinking as
$M/N$ grows \citep{marchenko1967,golub2013matrix}, and a raw monomial dictionary
at $d \ge 3$ over variables of mixed scale is a Vandermonde-like matrix whose
condition number
can exceed the noise floor, at which point the ``nullspace'' is numerical
fiction.

\begin{table}[t]
\caption{Monomial dictionary size $M = \binom{n+d}{d}$. The support search
scales with $M$, not with $n$; degree is as important as dimension. A
degree-$1$ (linear) invariant class such as stoichiometric conservation is
tractable at any $n$ encountered in practice.}
\label{tab:dict}
\begin{center}
\begin{tabular}{r|rrrr}
$n$ & $d{=}1$ & $d{=}2$ & $d{=}3$ & $d{=}4$ \\
\hline
$4$  & $5$  & $15$  & $35$   & $70$   \\
$6$  & $7$  & $28$  & $84$   & $210$  \\
$8$  & $9$  & $45$  & $165$  & $495$  \\
$11$ & $12$ & $78$  & $364$  & $1365$ \\
$17$ & $18$ & $171$ & $1140$ & $5985$ \\
\end{tabular}
\end{center}
\end{table}

We address each in turn.

\subsection{Gr\"obner deflation}

Section~IV-A of the main paper gives the argument: orthogonal projection removes
$\mathrm{span}(g)$, one direction, where the trivial content is the whole
filtered piece $\langle g\rangle \cap \mathcal{P}_d = g \cdot
\mathcal{P}_{d - \deg g}$. Table~\ref{tab:lin-vs-gb} counts the divergence,
which is invisible at the degree where a quadratic invariant is first recovered
and decisive as soon as the adaptive loop lifts degree.

The shortfall is worse than one direction per generator, in a way that matters
once more than one invariant has been found. Projecting a second
generator $g_2$ out of a subspace from which $g_1$ has already been projected
removes nothing at all whenever $\langle g_1, g_2\rangle \ne 0$ as coefficient
vectors. This is because, after the first step, the subspace is
$S \cap g_1^\perp$ (exactly, since $g_1 \in S$), which no longer contains
$g_2$, and $v \mapsto v - \langle v, g_2\rangle g_2/\|g_2\|^2$ has kernel
$\mathrm{span}(g_2)$, so it is injective on any subspace meeting that span
trivially and the rank does not drop. The inner product here is the one on
monomial coefficient vectors; the shortfall is therefore basis-dependent, where
normal-form reduction is not, which is a further reason to prefer it. Acrobot's
two unit-norm
generators both carry the constant term $-1$, so their coefficient vectors have
inner product $1$ and this is exactly the case that occurs: sequential
projection removes one direction in total, not two, and we measure precisely
that (Table~\ref{tab:deflation}, and Appendix~\ref{app:deflation} for the
orthogonal-generator case where the shortfall does not appear).

\begin{table}[t]
\caption{Deflating a single degree-$2$ invariant $g$ out of the degree-$\le d$
nullspace, $n = 4$. Orthogonal projection removes one direction at every
degree; normal-form reduction removes the entire filtered piece
$\langle g\rangle \cap \mathcal{P}_d$. The two agree only at $d = \deg g = 2$.}
\label{tab:lin-vs-gb}
\begin{center}
\begin{tabular}{cccc}
$d$ & multiples of $g$ present & projection removes & normal form removes \\
\hline
$2$ & $1$  & $1$ & $1$  \\
$3$ & $5$  & $1$ & $5$  \\
$4$ & $15$ & $1$ & $15$ \\
\end{tabular}
\end{center}
\end{table}

\begin{table}[t]
\caption{Nullspace deflation on Acrobot-v1, degree $\le 3$, $n=6$, $M=84$
monomials, single-energy passive trajectory, measured on sampled data and not
on the exact ideal. The $14$ trivial multiples vanish to machine precision
at every step size, being exact algebra; the energy direction vanishes only to
the integrator's drift floor, so admitting it requires a nullspace tolerance
above that floor. Sequential orthogonal projection of $g_1$ then $g_2$ removes
one direction in total, not two, for the reason given in
Section~IV-A of the main paper. The last column is the residual of the single
surviving direction outside $\mathrm{span}\{E, 1\}$, and it is the column that
matters: at $\Delta t = 0.2$ the count collapses to $1$ just as it does at
$\Delta t = 0.002$, but the survivor is not the energy.}
\label{tab:deflation}
\begin{center}
\small
\resizebox{\textwidth}{!}{%
\begin{tabular}{cc|ccc|cl}
$\Delta t$ & tol & raw $d_{\mathrm{null}}$ & after projection
  & after normal form & residual & outcome \\
\hline
$0.2$   & $10^{-6}$  & $14$ & $13$ & $\mathbf{0}$ & --
  & energy below tolerance \\
$0.2$   & $10^{-4}$  & $15$ & $14$ & $\mathbf{1}$ & $6.2\times10^{-1}$
  & survivor is \emph{not} the energy \\
$0.05$  & $10^{-4}$  & $16$ & $15$ & $2$ & --
  & spurious direction admitted \\
$0.01$  & $10^{-6}$  & $15$ & $14$ & $\mathbf{1}$ & $1.6\times10^{-7}$
  & energy isolated exactly \\
$0.002$ & $10^{-6}$  & $15$ & $14$ & $\mathbf{1}$ & $4.9\times10^{-10}$
  & energy isolated exactly \\
\end{tabular}%
}
\end{center}
\end{table}

At $d_{\mathrm{null}} = 15$ with one informative direction the condition is
close to vacuous. Reducing every candidate to its normal form modulo the
reduced Gr\"obner basis of what has already been found removes the trivial
directions exactly, by the ideal-membership criterion
\citep[Chapter 2, Section 6]{cox2015ideals}: $q \in \langle G \rangle$ if and only if
$\NF_G(q) = 0$.

\paragraph{The dimension count alone does not certify recovery.}
Table~\ref{tab:deflation} makes this concrete. At Gymnasium's default
$\Delta t = 0.2$ the sequence $15 \to 1$ occurs exactly as at
$\Delta t = 0.002$ and the arithmetic looks identical, but $62\%$ of the
surviving direction lies outside $\mathrm{span}\{E, 1\}$: the tolerance needed
to see past the integrator's drift floor at that step size is loose enough to
admit a spurious direction, and deflation faithfully removes the trivial
content around it. Acrobot's energy is not recoverable at the default timestep.
Deflation is what makes the surviving list short enough to audit, and it is the
residual check, not the count, that decides whether the survivor is the
invariant. On a system with no ground truth that check is unavailable, which is
the practical cost of the tension recorded in Remark~\ref{rem:tension}.

\paragraph{Implementation.} Reducing each floating-point nullspace vector
individually is numerically poor: the reduced set is not orthonormal and
near-zero survivors are awkward to threshold. We instead build the matrix of the
linear map $\NF_G$ restricted to the degree-$\le d$ monomial space, whose
entries are exact rationals, apply it to the nullspace basis and
re-orthonormalise by SVD. The SVD is
taken on the column-normalised design matrix, so its right singular vectors are
coordinates against normalised columns and must be divided back by the column
norms before any algebraic operation touches them. Reducing in the wrong
coordinates leaves the nullspace dimension unchanged, which presents as
deflation doing nothing at all.

\subsection{Coordinate lifting}
\label{sec:lifting}

Non-polynomial invariants can be brought into scope by adjoining variables
together with their defining relations: $c = \cos\theta,\, s = \sin\theta$ with
$c^2 + s^2 - 1$; $r$ with $r^2 - (x^2 + y^2)$ to clear $1/r$ potentials; unit
quaternions for spatial attitude. This does more than enlarge the
hypothesis class: it is the difference between a system having a polynomial
invariant and not having one: in raw joint coordinates the kinetic energy
$\tfrac12 \dot q^\top M(q) \dot q$ of an articulated body is transcendental,
because $M(q)$ is built from trigonometric functions, whereas in
$(\cos, \sin)$-lifted coordinates it is polynomial, using
$\cos(\theta_1 \pm \theta_2) = c_1c_2 \mp s_1s_2$.

Lifting increases $n$. It also, counterintuitively, tends to make the search
\emph{easier}, because the adjoined relations are themselves low-degree
generators involving few variables that are deflated against by our normal-form reduction
procedure, removing exactly the directions that would otherwise clutter the
higher-degree nullspace (Table~\ref{tab:deflation}).

\subsection{Conditioning and units}
\label{sec:units}

Two preprocessing steps are not optional.

\paragraph{Orthogonal dictionary for the numerical stage.} Standardise each
variable over the sampled data, build the dictionary from tensorised Legendre
or Chebyshev polynomials, not raw monomials
\citep{trefethen2019approximation}, perform nullspace extraction and support
search there, then convert \emph{exactly} back to the monomial basis, the
change of basis being an exact triangular rational matrix, before snap-rounding
and Gr\"obner canonicalisation. The exactness of the algebraic
stage is preserved while the numerical stage is well conditioned.

\paragraph{Nondimensionalisation before snap-rounding.} The rationality prior,
with its denominator ceiling $Q_{\max}$, is a prior on the \emph{units}, not on
the physics. Reacher's forward-kinematics identity has coefficients
$0.1 = 1/10$ and $0.11 = 11/100$ in metres; at $Q_{\max}=16$ the second is
unrepresentable and the invariant is not recoverable. Rescaling lengths by the
first link length gives $1$ and $11/10$, both admissible
(Appendix~\ref{app:reacher}); the ratio $0.11/0.1$ is computed in floating
point, $1.0999999999999999$ in IEEE arithmetic rather than $1.1$, and is
cleaned up by an uncapped rational reconstruction rather than by exact
arithmetic throughout, unambiguously in this case. We therefore
nondimensionalise by characteristic
length, time and mass \citep[cf.][]{udrescu2020aifeynman} and normalise each
candidate by its largest coefficient before rounding. For coefficients that are
genuinely irrational, as mass and inertia \emph{ratios} of real hardware usually
are, an integer-relation method such as PSLQ \citep{ferguson1992pslq} would
provide a fallback, but is not used here, since the reference systems used
have rational coefficients.

\section{What Canonicity Provides: Extended Discussion}
\label{app:canonicity}

Canonicalisation is not free. It requires exact rational arithmetic, a
snap-rounding stage that can fail, and a Gr\"obner computation whose worst case
is doubly exponential. What it provides is a short list, and that list determines
what the rest of
the paper does.

\paragraph{Provided: decidable equality.} Section~V of the main paper states it
and Appendix~\ref{app:related} contrasts it with a border basis. One practical
consequence follows: when one side is already available, as a reference
basis typically is, no recomputation is needed at all, since by the
ideal-membership criterion \citep[Chapter 2, Section 6]{cox2015ideals}
$q \in \langle G\rangle$ exactly when $\NF_G(q) = 0$, so containment in both
directions decides equality by division alone.

\paragraph{The decision, run.} Three configurations, in increasing distance from the
specification.

\emph{Declared.} The test system's ideal is built from the physics its fault
declares, so the decision has a known answer. Parameter faults change a constant
and move the ideal; observation faults are a change of coordinates on the same
physics, so their ideal is the image of the reference set under the inverse of
that change, and it also differs; a different integrator or step size leaves the
algebra untouched, so the ideal is identical; and a fault that destroys
conservation leaves no ideal of this form at all, which is a third answer rather
than a gap. All $19$ decisions agree with the declaration, the nineteen being the
fifteen catalogue faults, the two healthy controls and the two systems built from
installed constants below, and the third and
fourth cases are what no residual can separate, since a coarser integrator and a
wrong mass both raise the residual and only one of them moves the ideal.

\emph{Recovered.} The test system's ideal is built from trajectories. Recovery
returns a numerical direction, attribution fits the parameter it implies, and
the fitted value is rationalised to the nearest rational of denominator at most
$1000$ before the exact comparison, which is the step where the prior that a
specification holds round numbers re-enters. This ceiling is separate from
$Q_{\max}=16$ of Section~\ref{sec:units}: it bounds a single fitted value
rather than a discovered coefficient vector, is set for a different reason,
and is the only denominator cap actually enforced anywhere in the pipeline
reported here. The comparison is available exactly
where attribution names a parameter, $7$ of the $19$, and on those seven it
reproduces the declared ideal exactly, recovering $13/10$, $6/5$, $3/5$, $3/2$,
$981/100$, $13/10$ and $9/10$. Where attribution declines there is no second
ideal to compare and the entry is unavailable, not equal; reporting
agreement on the strength of having found nothing would invert what the screen
has already said about those systems.

\emph{Installed.} The test system's ideal is built from the constants the
installed simulator carries. Acrobot's masses, lengths, centres of mass and
inertias are read from the class attributes of the imported
\texttt{AcrobotEnv}, and the resulting ideal is equal to the reference ideal and
unequal to the $m_2 = 1.3$ fault's. Reacher's link lengths are read from the
compiled MuJoCo model as the body-frame offsets of \texttt{body1} and
\texttt{fingertip}, $0.1$ and $0.11$, and that ideal is equal to the reference
too. Reading the lengths from the two capsule geoms instead gives $0.1$ for
both, since the second link's geom stops short of the fingertip body, and the
ideal built from those is unequal. That is a property of the shipped model file
and the exact decision is what makes it visible.

\paragraph{Not needed: the screen.} Section~V of the main paper reports the
ablation and Table~\ref{tab:approx} carries it in full. Two ways of being
approximate are separated there, because they fail through different mechanisms.

\emph{Panel (a): perturbed coefficients.} Every coefficient of both reference
sets is moved by a
relative $\pm\epsilon$, with the sign drawn per coefficient and $\epsilon$ an
exact power of ten, so the generators still span the right directions and only
the numbers are wrong. A uniform sign would rescale a generator, which is gauge:
the screen's residual is scale-relative and the ideal is unchanged. The axis of
panel (a) is therefore the measured deviation after a scale is fitted out, which
sits a little under the nominal $\epsilon$. Keeping the perturbation in
$\mathbb{Q}$ is what lets the same set be handed to the equality decision, so the
last column of that panel is a property of the object the first four columns
were measured on.

\emph{Panel (b): a set recovered from data.} The generating set is read off
singular directions of the healthy reference system, unsnapped and
uncanonicalised, at the observation noise the screen then runs at. Exactness is granted where the pipeline is not what is
under test: the quotient projection inside recovery needs a basis over
$\mathbb{Q}$ and is given the exact unit-norm identities, which are relations
between an angle's sine and cosine that no fit produced. What stays approximate
is every coefficient the screen evaluates. A generator the recovery abstains on
is dropped, which lowers the Bonferroni count to match the hypotheses actually
tested, and the fault living in it is scored as a miss. The \emph{forced} rows
answer the obvious objection, that a pipeline willing to accept a weaker
spectral gap would have kept it: it would, and at $\sigma = 10^{-4}$ the
generator it keeps has a gauge-fitted coefficient deviation of $1.00$ from the
energy, so the screen then reports a broken constraint under a name that no
longer denotes it. Panel (b) reports both.

\begin{table}[t]
\caption{An approximate reference set in place of the exact one, over the
fifteen catalogue faults and two healthy controls. \emph{(a)} The exact set with
every coefficient perturbed, at $\sigma = 0$; ranges are over five sign patterns
and both regimes, and the last column is the exact ideal-equality decision on the
same perturbed set. \emph{(b)} The set a numerical recovery returns from healthy
reference data, screened at the observation noise it was recovered at, against
the exact set on the same data; \emph{forced} removes the spectral-gap test so a
generator is returned whatever the conditioning. Deviations are relative and
gauge-fitted.}
\label{tab:approx}
\centering
\small
\begin{tabular}{lcccc}
\multicolumn{5}{l}{\emph{(a) Perturbed coefficients, $\sigma = 0$}} \\
\toprule
measured deviation & detected & localised & false alarms
  & $\langle G_\epsilon\rangle = \langle G\rangle$ \\
\midrule
$0$                                 & $15/15$    & $15/15$    & $0/4$   & yes \\
$\le 9.8\times10^{-5}$ (6 decades)  & $14$--$15/15$ & $14$--$15/15$ & $0/120$ & no  \\
$9.8\times10^{-4}$                  & $13$--$14/15$ & $13$--$14/15$ & $0/20$ & no \\
$9.9\times10^{-3}$                  & $9$--$13/15$  & $9$--$13/15$  & $0/20$ & no \\
$1.0\times10^{-1}$                  & $4$--$8/15$   & $3$--$5/15$   & $0/20$ & no \\
\bottomrule
\end{tabular}

\vspace{1.2ex}

\begin{tabular}{llcccc}
\multicolumn{6}{l}{\emph{(b) A recovered set against the exact one, at matched
observation noise}} \\
\toprule
reference set & $\sigma$ & generators & worst deviation & detected & localised \\
\midrule
exact     & $0$ & $8/8$ & $0$                  & $15/15$ & $15/15$ \\
recovered & $0$ & $8/8$ & $2.4\times10^{-8}$   & $15/15$ & $15/15$ \\
forced    & $0$ & $8/8$ & $2.4\times10^{-8}$   & $15/15$ & $15/15$ \\
\addlinespace
exact     & $10^{-8}$ & $8/8$ & $0$                & $14/15$ & $14/15$ \\
recovered & $10^{-8}$ & $7/8$ & $1.0\times10^{-5}$ & $14/15$ & $14/15$ \\
forced    & $10^{-8}$ & $8/8$ & $1.0\times10^{-5}$ & $15/15$ & $14/15$ \\
\addlinespace
exact     & $10^{-6}$ & $8/8$ & $0$                & $14/15$ & $14/15$ \\
recovered & $10^{-6}$ & $7/8$ & $7.0\times10^{-4}$ & $14/15$ & $14/15$ \\
forced    & $10^{-6}$ & $8/8$ & $7.0\times10^{-4}$ & $15/15$ & $14/15$ \\
\addlinespace
exact     & $10^{-4}$ & $8/8$ & $0$                & $13/15$ & $13/15$ \\
recovered & $10^{-4}$ & $6/8$ & $4.4\times10^{-4}$ & $5/15$  & $3/15$  \\
forced    & $10^{-4}$ & $8/8$ & $1.00$             & $12/15$ & $11/15$ \\
\bottomrule
\end{tabular}
\end{table}

Three readings follow. The screen tolerates four decades more coefficient error
than a numerical recovery on clean data incurs, so on that evidence exactness is
not a precondition for localisation. No perturbation magnitude anywhere in panel
(a) raised a false alarm, over $184$ healthy controls, so a wrong reference set
loses faults without manufacturing them, which is the safer of the two ways to
be wrong. And the failure that does arrive in panel (b) is one of availability:
past $\sigma = 10^{-6}$ the recovery cannot produce the energy generator at all,
where a specification supplies it at any noise level, which is the sense in which
the reference set and the pipeline that might have found it are separate assets.

The equality decision has the opposite profile. It separates $\langle G\rangle$
from a copy perturbed at $10^{-12}$, in all $90$ draws, eight decades below the
coefficient error the screen absorbs without losing a fault. Run on a set
recovered from data, it would therefore report a healthy system as different
every time. The decision is exact or it is nothing.

\paragraph{Not provided: minimality.} Remark~\ref{rem:notminimal} gives the
twisted-cubic counterexample and its consequence for a loss of the form
$\sum_{p \in G}\|p(\cdot)\|^2$. Our Reacher run reproduces the point unprompted:
the recovered basis has ten elements for an ideal that five invariants generate.

\paragraph{Not provided: safety.} Section~V of the main paper states why. It is
tempting to argue otherwise via the step ``if $a$ is safe then
$p_i(s'_{\rm true}) = 0$'', which conflates safety with membership of the
constraint variety. The most invariants can do for a runtime shield is shrink
the uncertainty set it reasons over, replacing a query about the ball
$B(\hat s',\epsilon)$ with one about $B(\hat s',\epsilon) \cap \AVar(I)$, which
is strictly smaller and therefore less conservative at equal safety. That
benefit is real but modest, it requires a safety specification supplied from
outside, and it does not require canonicity, since any correct generating set
shrinks the same set by the same amount. We note it and do not pursue it.

\begin{remark}[The tension between tolerating drift and resolving it]
\label{rem:tension}
An invariant-based shield must tolerate its integrator's drift, and a
drift diagnostic exists to resolve it, so the two pull in opposite directions.
On Acrobot at Gymnasium's $\Delta t = 0.2$ with RK4, relative energy drift over
a $40\,$s passive rollout is $6.5\times10^{-4}$, falling to $6.9\times10^{-7}$
at $\Delta t = 0.05$ and $2.2\times10^{-10}$ at $\Delta t = 0.01$
(Appendix~\ref{app:drift}). The same floor bounds \emph{discovery}: the energy
direction enters the numerical nullspace only once
the tolerance exceeds the drift, and a tolerance that loose already admits
spurious content. The usable window closes as $\Delta t$ coarsens, and on
Acrobot it has closed by the shipped step size. Discovering the conserved
quantity therefore requires re-integrating the environment more finely than it
ships. Section~VII-C of the main paper runs into this directly: the potential must be
discovered on a re-integrated copy and then deployed on the environment as
distributed.
\end{remark}

The conclusion we draw is that these diagnostic uses are the ones to build,
and the rest of the paper builds them.

\section{Construction of the Two-Level Diagnostic}
\label{app:diagnostic}

Section~VI of the main paper fixes the input and the two levels. This section
carries the construction of each.

\subsection{Recovery in the quotient: construction and gap sensitivity}

Section~IV-B of the main paper gives the construction, writing $\mathcal{P}_d$ for
the polynomials of degree at most $d$, $G_{\rm known}$ for what has already been
established, and $\Pi$ for the projection induced by the normal-form map. This
section carries what the body defers.

The quotient removes the \emph{absolute} tolerance, since no nullspace dimension
is decided and no singular value is compared against a number carrying the units
of the dictionary. It does not remove a decision. Acceptance is by the
dimensionless ratio $\sigma_{\min}/\sigma_{\text{next}} < \gamma$ with
$\gamma = 0.1$ throughout, and the noise ceiling of Table~\ref{tab:attrnoise} is
partly a statement about $\gamma$. On the $m_2$ fault at lag $50$ the measured
gap rises from $3.0\times10^{-4}$ at $\sigma = 10^{-8}$ through
$3.0\times10^{-3}$ at $10^{-7}$ to $2.9\times10^{-2}$ at $10^{-6}$ and $0.29$ at
$10^{-5}$, so the verdict at $10^{-5}$ is set by $\gamma$ and not by the fit,
which is never reached. Over the range where a fit does run, its residual moves
in the same direction, from $9.1\times10^{-6}$ at $\sigma = 10^{-8}$ to
$7.7\times10^{-4}$ at $10^{-6}$. The ceiling is therefore not an artefact of
$\gamma$ alone, though it is not independent of it either.

\paragraph{Where the orthonormality has to hold.} $\Phi$ is column-normalised
before its decomposition, so that one gap threshold means the same thing across
a dictionary whose columns span many orders of magnitude. The quotient basis is
built in monomial coordinates and therefore has to be carried into the
normalised ones, and a column-by-column rescaling of an orthonormal basis is not
orthonormal. Without re-orthonormalising it there the
singular values of $\Phi\Pi^\top$ are those of a composition with a
non-orthogonal map, not those of $\Phi$ restricted to the subspace, and
the accepted ratio then varies with the units of the state variables: scaling a
single velocity coordinate by $100$ moves it by an order of magnitude, whereas the
orthonormalised ratio is constant to seven digits.

Two measurements sit behind Section~IV-B of the main paper. First, deflation is a
precondition here rather than an
efficiency measure. Run on Acrobot without the projection, the least-varying
direction of $\Phi$ lies \emph{entirely} inside the $14$-dimensional trivial
component, to a measured fraction of $1.0000$, and has alignment $0.0000$ with
the energy; run with it, the alignment is $1.0000$
(Appendix~\ref{app:verified}). Trivial multiples are conserved for algebraic
reasons and say nothing about the physics, so without the quotient there is not
a harder search to do, there is nothing to find. Second, the lag $k$ is free
robustness. A conserved quantity has zero difference at every lag, so
lengthening the lag leaves the signal where it was while every non-conserved
direction grows with elapsed time and the measurement noise does not grow at
all. The recovery depends on a ratio between those, so it improves roughly in
proportion to $k$, until secular integrator drift over the longer window begins
to matter. Empirically $k = 50$ at $\Delta t = 0.005$ provides about a decade of
noise tolerance (Table~\ref{tab:attrnoise}).

\paragraph{Snap-rounding is deliberately bypassed.} Snap-rounding is a heuristic that recovers exact rational coefficients by rounding numerical values to nearby small fractions. It assumes the system is correct (coefficients are round numbers such as 
0.1
0.1 or 
0.11
0.11). A faulty system, however, carries no such guarantee; its invariants generically have coefficients that are not small rationals. Snapping them would either fail outright or, worse, round a faulty coefficient back onto the healthy value and hide the fault. The diagnostic therefore works with the raw numerical direction and lets the template fit of Section~VI-B of the main paper supply the structure that snap-rounding would otherwise have supplied.

\subsection{Level 1 screening: construction and calibration}

Section~VI-A of the main paper gives the test. The residual it ranks is, for a
vanishing generator, the scale-relative quantity
\begin{equation}
\label{eq:relresid}
\rho(p; X) \;=\;
  \frac{\big(\tfrac{1}{N}\sum_i p(x_i)^2\big)^{1/2}}
       {\big(\tfrac{1}{N}\sum_i (\sum_\alpha
|c_\alpha|\,|x_i^\alpha|)^2\big)^{1/2}},
\end{equation}
whose denominator is the natural scale of $p$ on that data, and for a conserved
generator the same expression with the numerator replaced by the standard
deviation of $p$ \emph{along} each trajectory. What follows is what the body
defers: the absolute floor the rank test forces, what the reported $p$-values do
and do not carry, and the measured false-alarm rate.

\paragraph{An absolute residual floor, and why one is unavoidable.} The rank
test has no scale. Given $32$ blocks per configuration whose residuals differ
systematically in the seventeenth significant figure, it separates them
completely and returns $p \approx 3\times10^{-12}$, reporting a broken
constraint on a system exact to machine precision. Some absolute floor is
therefore forced, and the only question is whether it is stated. We set it at
$\rho_{\min} = 10^{-8}$ and declare two configurations indistinguishable, without running
the test, when no residual on either configuration rises above it. The rule is absolute
and nothing else; a relative tolerance alongside it would be a second and
undeclared threshold, suppressing any fault whose residual ratio happens to lie
close to one whatever the absolute size of the residuals.

The floor applies to the residual in the generator's units and not to the
scale-relative quantity the rank test ranks, and the two are not
interchangeable. Where a generator's scale is carried by a single term, the term
scale in \eqref{eq:relresid} is the root-mean-square of that same term, so the
ratio is identically $1$ whenever the term is non-zero and $0$ otherwise and
carries no magnitude at all. Applied to that ratio the floor is unreachable, and
on Reacher's $d_z$ generator a $z$-leak of $10^{-30}$ then screens as a broken
physical constraint with the same $p$-value as a leak of $10^{-3}$. The value is
a measurement: shipped classic-control observations are float32, which floors
the unit-norm identities at $1.7\times10^{-8}$ however exact the dynamics, and
shipped MuJoCo observations floor at about $10^{-8}$ for the reason
Section~VII-D of the main paper gives. Removing the floor raises the measured false-alarm
rate on Acrobot from zero in $500$ null comparisons to $10.2\%$
(Fig.~3 of the main paper). The cost is a blind spot: a genuine violation
smaller than $\rho_{\min}$ is invisible to the screen, and the Reacher finding
of Section~VII-D of the main paper, at $6\times10^{-9}$, sits inside it and was found by
the exact stage. This does not reintroduce the fixed threshold argued against
above, which was a threshold on the \emph{fault} signal; $\rho_{\min}$ is a
floor on the \emph{instrument}, set by the observation dtype and not by the
expected size of a fault.

\paragraph{What the $p$-values do and do not carry.} With $32$ blocks per configuration
the smallest attainable one-sided $p$ is $3.26\times10^{-12}$, reached whenever
every test block exceeds every reference block. Every fault the screen detects
at $\sigma = 0$ attains it, so the reported $p$ is the test's resolution limit
and not a graded measure of evidence: a $30\%$ mass error and a $0.1\%$ error in
$g$ both return $9.77\times10^{-12}$ after the three-generator correction. The
$p$-value should therefore be read as a decision at a controlled level, and the
residual \emph{ratio} $\rho_{\rm test}/\rho_{\rm ref}$, which we report
alongside it, as the effect size. The ratio does discriminate where the
$p$-value does not: on Acrobot it runs from $1.19\times10^{7}$ for the $0.1\%$
error in $g$ to $2.08\times10^{9}$ for the $30\%$ mass error. The MMD baseline
carries the same limit from the other side, its permutation $p$ flooring at
$1/(B+1)$ for $B$ permutations, which is why $B$ is set so that the floor lies
two orders below $\alpha$, not beside it.

\paragraph{Blocks are not independent, and the rate says so.} The blocks are
contiguous stretches of the same trajectories, four per trajectory over eight,
so blocks sharing a trajectory share an initial condition and hence an energy
level and a drift rate. For a conserved generator they are positively
correlated, and the rank test, counting them as independent replicates, returns
a $p$-value smaller than the design supports. The alternative is one residual
per trajectory with an exact permutation null over the $\binom{16}{8} = 12870$
splits, whose smallest attainable $p$ is $7.8\times10^{-5}$ before correction,
still an order inside $\alpha$. Fig.~3 of the main paper measures the size of the
difference.

The block-level test on Acrobot exceeds the nominal level in three of its four
settings, at $10.2\%$ against a nominal $1\%$ once the residual floor is removed
and at $2.6\%$ under measurement noise with the floor in place or without it;
all three intervals lie entirely above $\alpha$, and in each the excess is
carried mainly by the energy generator, whose residual is a drift over a window
and whose blocks are therefore the most strongly coupled. The trajectory-level
test is consistent with nominal throughout, raising nothing at all in three of
the eight settings, its worst rate being $1.4\%$ $[0.7, 2.9]$ on noisy Acrobot,
whose interval covers $\alpha$. Reacher, sampled as independently drawn poses
and not as stretches of a trajectory, keeps a rate at or below $\alpha$ under
both units, its worst being $0.8\%$ $[0.3, 2.0]$; three of its four intervals
cover $\alpha$, so the difference between the two units is not resolved there
and only the Acrobot settings separate them.

So the stated level is supported at the trajectory level and not at the block
level, and since the trajectory-level test is also the more powerful of the two
under noise (Table~I of the main paper), there is no reason to prefer blocks. Two
further readings separate. At $\sigma = 0$ with the floor in place both units
return zero in $500$ trials, an interval of $[0, 0.008]$ strictly inside
$\alpha$, so the noiseless claim is a measured rate. And the floor does the work
it was introduced for only at zero noise: removing it changes the block-level
Acrobot rate from $0.000$ to $0.102$ there, and changes nothing at
$\sigma = 10^{-3}$, where the residuals have risen off the floor and the
dependence between blocks is what remains.

The calibration has one limit. The Bonferroni correction is applied across the
generators of one system pair, with no correction across the fifteen faults.
This is the right choice for an instrument pointed at one system at a time,
and the wrong one for reading the catalogue as a single experiment.

\subsection{Level 2 attribution: construction and verdict logic}

Let $p(x;\theta)$ be a template whose coefficients depend symbolically on
physical parameters $\theta$, for instance the forward-kinematics identity
carrying $\ell_1$ and $\ell_2$, or the Acrobot energy carrying masses, lengths
and inertias, and let $\hat p$ be the polynomial recovered from the test system.
Recovery returns a \emph{direction}, so $\hat p$ arrives with an arbitrary scale
and, for a conserved quantity, an arbitrary additive constant as well, since $E$
and $\alpha E + \beta$ are conserved together. Writing $c_T(\theta)$ and $c_R$
for the coefficient vectors over the union of the two supports, we fit modulo
that gauge by minimising
\begin{equation}
\label{eq:tmplfit}
r(\theta) \;=\;
\min_{\alpha,\beta}\;
\frac{\big\| c_T(\theta) - \alpha\, c_R - \beta\, e_0 \big\|_2}
     {\big\| c_T(\theta) \big\|_2},
\end{equation}
with $e_0$ the constant coordinate, dropped for vanishing generators. The
normalisation makes $r$ dimensionless, so residuals from different hypotheses
are comparable and the ranking over single-parameter hypotheses is meaningful.
A template carries more monomials than parameters, which is why a wrong
hypothesis shows up as a large residual instead of an effortless fit.

Section~VI-B of the main paper states the three verdicts. The one that needs
elaborating is \emph{not explicable as a parameter fault}, which is reached in
two ways: recovery finds no direction with a decisive spectral gap, or a
direction is recovered and no single-parameter hypothesis fits it. The first is
the correct verdict for a dropped Coriolis term, where energy is not conserved
under \emph{any} setting of the masses, so no parameter assignment explains the
observation and the diagnostic says so instead of returning the closest
available fit. The second reaches the same conclusion one stage later.

The two conditions on the winner are separate tests carrying separate verdicts,
and collapsing them is a mistake with a direction. The winner must fit
absolutely,
which rules out every hypothesis being bad and one merely least bad, and it must
beat the runner-up by a margin, which rules out two parameters entering the
generator in the same way and being genuinely indistinguishable from it. Failing
the second is non-identifiability, which is consistent with the specification and
is reported rather than resolved by tie-break. Failing the first is the opposite:
the recovered generator contradicts every hypothesis the template can express,
the specification included. Routing both to \emph{consistent with specification}
gives a system that contradicts its own specification a clean bill of health, and
on a $30\%$ mass error at $\sigma = 10^{-7}$ it does exactly that, with the
correct parameter ranked first at $1.3056$ and separated from its rival by a
factor of $10.9$, failing only the absolute tolerance and by a factor of $1.4$.

\subsection{The two downstream constructions in full}

The same generators support two constructions that the introduction claims do
\emph{not} require exactness. We define them here and test that claim in
Sections~VII-B and VII-C of the main paper.

\paragraph{Consistency regularisation.} Given $G = \{p_1,\dots,p_k\}$, train
$f_\theta$ with
\begin{equation}
\label{eq:lphys}
L \;=\; L_{\mathrm{MSE}} \;+\; \lambda \sum_{i} w_i\,
  \big\| p_i\big(f_\theta(s,a)\big) \big\|^2 ,
\end{equation}
penalising departure from the constraint manifold. For a conserved generator
the term is instead $\|p_i(f_\theta(s,a)) - p_i(s)\|^2$, the passive case of the
balance law \eqref{eq:balance}. The weights $w_i$ are not uniform and cannot be
made so by canonicalisation: by the twisted-cubic argument the basis may contain
algebraic consequences of its own elements, so uniform weighting over-penalises
whichever constraint generates the redundancy. We normalise each generator by
its scale on the training data, so a single $\lambda$ means the
same thing for a unit-norm constraint and for an energy of magnitude ten.

\paragraph{Potential-based shaping.} The obvious construction, an intrinsic
reward $-\sum_p \|p(s') - p(s)\|^2$ penalising induced invariant residual, is
identically zero in any real environment: the physics is never violated, so the
agent receives no signal whatsoever. The useful information is not that energy
is conserved but what its value is. For a discovered conserved quantity $E$ and
a task-determined target $E^\star$ we set $\Phi(s) = -|E(s) - E^\star|$ and
shape
\begin{equation}
\label{eq:shaping}
r_{\mathrm{shaped}}(s,a,s') \;=\; r_{\mathrm{ext}}(s,a,s')
  \;+\; \gamma\,\Phi(s') - \Phi(s),
\end{equation}
which by \citet{ng1999shaping} leaves the set of optimal policies unchanged, a
guarantee the residual formulation could not offer. On swing-up this recovers
from data the energy-pumping shaping that control engineers hand-design.

\section{Theoretical Analysis and Proofs}
\label{app:theory}

\subsection{What a zero constraint residual does and does not certify}
\label{sec:worldmodel-theory}

\begin{proposition}[Soundness and the necessity/sufficiency gap]
\label{prop:wm}
Let $G = \{p_1,\dots,p_k\}$ be \emph{vanishing} generators, generating an ideal
$I$ contained in the true vanishing ideal $I(\mathcal M)$ of the admissible
state manifold. Then:
\begin{enumerate}[label=(\alph*)]
\item \textbf{(Soundness.)} With weights $w_i > 0$, the penalty
  $\sum_i w_i \|p_i(\hat s')\|^2$ vanishes at every
  $\hat s' \in \mathcal{M}$, in particular at $\hat s' = \Phi(s,a)$. A model
  whose prediction lands on the constraint manifold is never penalised, so the
  regulariser cannot penalise a correct prediction.
\item \textbf{(Detection.)} Conversely, with $w_i > 0$ the penalty is zero
  \emph{only} on $\AVar(I)$, so any departure from the constraint variety along
  a direction some $p_i$ sees is registered. The content is the ``only'': no
  cancellation between generators can hide a violation, since every term is a
  square with positive weight.
\item \textbf{(Necessity, not sufficiency.)} Write $\mathcal{T}$ for the
  local trajectory manifold at $s'_{\rm true} = \Phi(s,a)$, the set of states
  reachable from $s$ over the admissible action set in one step. If the local
  dimension of $\AVar(I)$ at $s'_{\rm true}$ exceeds $\dim \mathcal{T}$, then
  $L_{\mathrm{cons}} = 0$ does not imply the transition is dynamically valid:
  there are $\hat s' \in \AVar(I)$ arbitrarily close to $s'_{\rm true}$ and
  unreachable from $s$. Sufficiency requires the two to coincide locally, which
  does not follow from canonicalisation and must be tested
  (Section~\ref{sec:sufficiency}).
\item \textbf{(Not one penalty per constraint.)} The elements of a reduced
  Gr\"obner basis have pairwise non-dividing leading monomials
  \citep[Chapter 2, Section 7]{cox2015ideals}, but this does not make them a minimal
  generating set, so $L_{\mathrm{cons}}$ may include terms that are algebraic
  consequences of others. What canonicalisation does guarantee is that
  $L_{\mathrm{cons}}$ is a well-defined function of the ideal and the monomial
  order, hence reproducible across runs and across discovery procedures.
\end{enumerate}
The hypothesis that $G$ is vanishing does work here. A reference set such as
Acrobot's contains a conserved generator, and the energy does not vanish on
$\mathcal{M}$, so (a) and (b) do not apply to it as stated. For a conserved
generator the corresponding statement is about the transition graph and uses
the difference form $\|p(f_\theta(s,a)) - p(s)\|^2$, which is the form
Section~VI-C of the main paper actually trains against: that penalty vanishes on
every transition along which $p$ is conserved, and is nonzero exactly where $p$
moves. The necessity/sufficiency gap of (c) is unchanged either way.
\end{proposition}

\begin{proof}
(a) and (b) are immediate from $p_i \in I \subseteq I(\mathcal M)$ and the
definition of $\AVar(I)$. For (c), the hypothesis is on the \emph{local}
dimension at
$s'_{\rm true}$, since a global dimension bound says nothing about the
neighbourhood in question: a real algebraic set can carry a high-dimensional
component far away and be a single point nearby. Under that local hypothesis
every neighbourhood of $s'_{\rm true}$ in $\AVar(I)$ meets the complement of
$\mathcal T$, and any such point is a counterexample. For (d), the
non-dividing-leading-monomial property follows from reducedness: if
$\mathrm{LT}(p_i) \mid \mathrm{LT}(p_j)$ then $p_j$ would reduce against
$p_i$. That this does not imply minimality is witnessed by the twisted cubic
(Appendix~\ref{app:tc}), whose reduced basis has three elements while the
ideal has two generators.
\end{proof}

\subsection{Why screening localises and a two-sample test cannot}
\label{sec:whylocalise}

The asymmetry the experiments measure is structural, not a matter of
statistical power.

\begin{proposition}[Localisation]
\label{prop:localise}
Let $\mathcal{A}$ and $\mathcal{B}$ be two systems and let $p \in G_{\rm ref}$.
If $p$ vanishes identically on $\mathcal{A}$'s reachable set and
$\rho(p;\,X_\mathcal{B}) > 0$ for data $X_\mathcal{B}$ from $\mathcal{B}$, then
$\mathcal{B}$ violates the specific constraint $p$, and this conclusion is
invariant to any reparameterisation of the sampling distribution over
$\mathcal{B}$'s reachable set.
\end{proposition}

\begin{proof}
Immediate from \eqref{eq:relresid}: $\rho(p;X) > 0$ requires
$p(x_i) \neq 0$ for some sampled $x_i$, which exhibits a point of
$\mathcal{B}$'s reachable set off $\AVar(p)$. The conclusion is a statement
about the support of
$\mathcal{B}$'s reachable set, so no reweighting of the sampling measure can
invalidate it once such an $x_i$ has been exhibited.
\end{proof}

Two things bound how far this reaches. The proposition governs the exact
predicate $\rho > 0$, which in float64 is true of everything, whereas the
implemented screen of Section~VI-A of the main paper is a rank test on
scale-normalised block residuals and is itself a two-sample test of a different
statistic. What makes it insensitive to the sampling is that $\rho$ depends on
the state only through $p$ and is normalised by the term scale of $p$'s, so a
reseeded correct simulator moves the states without moving the residual; the
healthy controls of Section~VII-A of the main paper measure that, and the proposition does
not prove it. What the proposition establishes is the target the finite-sample
screen approximates: at the population level the question ``does $\mathcal{B}$
violate $p$'' is about the support of the reachable set and not the measure on
it. Sampling determines whether a violation is \emph{observed}, a measure
concentrated on $\AVar(p)$ yielding $\rho = 0$; what it cannot affect is the
conclusion drawn from an observed $\rho > 0$. None of this claims the screen is
more powerful, the discriminator detecting more faults at high noise. The claim
is that the two answer different questions.

\subsection{A computable sufficiency diagnostic}
\label{sec:sufficiency}

Proposition~\ref{prop:wm}(c) leaves open, for a given system, whether
sufficiency holds. Let $J_G(s) \in \R^{k \times n}$ be the Jacobian of
$(p_1,\dots,p_k)$ at a regular point $s$. By the regular-value form of the
implicit function theorem, the local level set of $G$ through $s$ has dimension
$n - \operatorname{rank} J_G(s)$. Estimate the local dimension
$d_{\mathrm{traj}}$ of the trajectory manifold independently. We supply it
analytically, which every system here permits. Local PCA over a neighbourhood is
the alternative where the dynamics are unknown, and on these systems it is not
accurate enough to carry the verdict: swept over threshold and neighbourhood
size it estimates a single Acrobot trajectory, analytically a curve, at between
$2$ and $6$, because curvature over a finite window contributes singular
directions that no threshold separates from genuine ones. Then
at generic sampled points, $\operatorname{rank} J_G(s) = n - d_{\mathrm{traj}}$
means the level set locally coincides with the trajectory manifold and
$L_{\mathrm{cons}} = 0$ is locally sufficient as well as necessary, while
$\operatorname{rank} J_G(s) < n - d_{\mathrm{traj}}$ means $G$ under-determines
the dynamics and only necessity holds. The same rank computation certifies
functional independence of conserved quantities
(Section~\ref{sec:algebra}), since $\nabla(E^2) \parallel \nabla E$.

This is local and sampling-dependent, not a global proof. The implicit function
theorem certifies equivalence only where the rank condition is checked, and
unsampled regions may have lower rank. On Acrobot the reference set has
$\operatorname{rank} J_G = 3$ against $n - d_{\mathrm{traj}} = 6 - 3$, so a zero
residual is locally sufficient at fixed energy; dropping the energy generator
leaves rank $2$ and the verdict changes to necessity alone. Note that
$d_{\mathrm{traj}}$ is a property of the sampling as much as of the system, since
an ensemble spanning many energy levels has a larger trajectory manifold than one
at fixed energy.

\section{Environments and the Fault Catalogue}
\label{app:environments}

The properties in this section are analytic or symbolic facts about the
environments, verifiable without training any agent; the verification code and
its output are given in Appendix~\ref{app:verified}. They are not experimental
results.

\subsection{Tier A: exact vanishing invariants, known ground truth}

Table~\ref{tab:tierA} lists the Tier-A environments together with their
exact invariants and the degree each requires.

\begin{table}[t]
\caption{Tier-A environments. All invariants listed are exact and hold under
arbitrary actuation unless marked \emph{(passive)}. The two environments in bold are the ones every
experiment in this paper runs on; the remaining four are stated to delimit the
class the construction applies to and carry no results here.}
\label{tab:tierA}
\begin{center}
\small
\begin{tabular}{lccl}
Environment & $n$ & deg & Invariants \\
\hline
Cartesian pendulum (DAE) & $4$ & $2$
  & $x^2{+}y^2{-}L^2$; $x\dot x{+}y\dot y$; energy \emph{(passive)} \\
\textbf{Acrobot-v1} & $6$ & $2$--$3$
  & $c_1^2{+}s_1^2{-}1$; $c_2^2{+}s_2^2{-}1$; energy \emph{(passive)};
    balance law \\
\textbf{Reacher-v4} & $\mathbf{11}$ & $1$--$2$
  & 2 unit-norm; $d_z{=}0$; 2 forward-kinematics \\
Free rigid body / attitude & $7$ & $2$
  & $\|q\|^2{-}1$; $\sum I_i\omega_i^2$, $\sum I_i^2\omega_i^2$
    \emph{(torque-free)} \\
Planar Kepler & $4$ & $2$
  & $x\dot y{-}y\dot x$; energy after lifting $r$ \\
Reaction network & $8$--$15$ & $\mathbf{1}$
  & moiety conservation, integer coefficients \\
\end{tabular}
\end{center}
\end{table}

Two rows carry most of the argument.

\paragraph{Acrobot-v1 is the worked example.} Its observation vector is already
$(c_1, s_1, c_2, s_2, \omega_1, \omega_2)$, and with the standard Gymnasium
parameters ($m_1{=}m_2{=}1$, $l_1{=}l_2{=}1$, $l_{c1}{=}l_{c2}{=}\tfrac12$,
$I_1{=}I_2{=}1$, $g{=}9.8$) the total mechanical energy is \emph{exactly} the
degree-$3$ polynomial
\begin{equation}
\label{eq:acrobotE}
E = \tfrac{7}{4}\omega_1^2 + \tfrac{1}{2}c_2\omega_1^2
  + \tfrac{5}{4}\omega_1\omega_2 + \tfrac{1}{2}c_2\omega_1\omega_2
  + \tfrac{5}{8}\omega_2^2
  - \tfrac{147}{10}c_1 - \tfrac{49}{10}c_1c_2 + \tfrac{49}{10}s_1s_2
\end{equation}
in that observation vector, agreeing with the coordinate-form energy to
$1.14\times10^{-13}$ over random states (Appendix~\ref{app:acrobot}). Every
coefficient denominator is at most $10$, comfortably inside a snap-rounding
ceiling of $Q_{\max}=16$ with no rescaling. Under torque
$\tau$ on the second joint, \eqref{eq:balance} specialises to
$E(s') - E(s) - \Delta t\,\tau\,\omega_2 = O(\Delta t^2)$, confirmed
numerically (Appendix~\ref{app:balance}). Acrobot thus exercises the vanishing
ideal, the conserved-quantity algebra, the balance law, deflation
(Table~\ref{tab:deflation}) and integrator drift (Remark~\ref{rem:tension}) in
one environment of six variables.

\paragraph{Reacher-v4 answers the dimensionality objection directly.} It is
eleven-dimensional, the same dimension as the locomotion tasks it replaces,
with observation
$(c_1,\allowbreak c_2,\allowbreak s_1,\allowbreak s_2,\allowbreak t_x,\allowbreak t_y,\allowbreak \omega_1,\allowbreak \omega_2,\allowbreak d_x,\allowbreak d_y,\allowbreak d_z)$, and it
satisfies five exact invariants of degree at most two: two unit-norm
constraints, $d_z = 0$, and the two forward-kinematics identities
\begin{equation}
\label{eq:reacherfk}
d_x + t_x - \ell_1 c_1 - \ell_2 (c_1c_2 - s_1s_2) = 0,
\qquad
d_y + t_y - \ell_1 s_1 - \ell_2 (s_1c_2 + c_1s_2) = 0,
\end{equation}
with residual $7.3\times10^{-17}$ over random poses
(Appendix~\ref{app:reacher}). Link lengths should be read from the environment
XML rather than assumed; with the standard $\ell_1 = 0.1$, $\ell_2 = 0.11$,
\eqref{eq:reacherfk} is recoverable only after nondimensionalisation
(Section~\ref{sec:units}), which makes it a direct test of that step.

\paragraph{Reaction networks show the ceiling is on degree, not dimension.}
Moiety conservation \citep{schuster1991conservation} is linear with integer
coefficients, so $M = n+1$ and the support search is trivial for any $n$ we
encounter. We run no reaction network here. The row is in the table because it
locates the ceiling analytically, on the degree and support size of the
invariant instead of on the number of state variables.

\subsection{Tier B: actuated and dissipative}

Actuated damped Acrobot, used to test recovery of the balance law
\eqref{eq:balance} in the joint $(s,a,s')$ space. The falsifiable prediction is
that the recovered damping coefficient matches the value in the environment
specification.

The system is the analytic Acrobot with viscous damping $b_1$, $b_2$ added at
the two joints and a torque drawn uniformly at each step, so it is both actuated
and dissipative and conserves nothing. Each transition is logged as
$(s, u, s')$ and nothing else is supplied: the recovery is not told the energy,
the damping, or that a balance law is what it should look for. The dictionary is
the union \eqref{eq:balance} requires, the difference dictionary in the six
observation coordinates joined to the degree-$2$ monomials of the velocities and
the torque scaled by $\Delta t$, and the difference block is projected onto the
quotient by the two unit-norm identities before the least-varying direction is
taken, for the same reason as in the passive case. The damping is then read from
the ratios between that direction's coefficients and the energy's own.

Across three damping settings and five step sizes from $\Delta t = 10^{-3}$ to
$2\times10^{-2}$, the recovered direction is decisive on $14$ of $15$
configurations and recovers both coefficients to a relative error of at most
$3.3\times10^{-3}$ on every one of them, for instance $b_1 = 0.10000$ and
$b_2 = 0.05000$ against specified $0.1$ and $0.05$. The prediction holds.

What the tier also shows is that the accuracy is the quadrature's and not the
recovery's. The left side of \eqref{eq:balance} is an exact difference where the
right side is an integral of power over the step, which a dictionary can carry
only as a quadrature. Under the first-order rectangle rule the direction clears
the spectral gap on $2$ of the same $15$ configurations and the damping it
yields is wrong by up to $96\%$. The balance law is recoverable only when the
dictionary discretises it to an order the recovery can resolve.

\subsection{Tier C: negative controls}
\label{sec:negative}

We retain Hopper and HalfCheetah \citep{todorov2012mujoco,towers2024gymnasium}
explicitly as \emph{negative controls}. These environments have no exact
polynomial invariant, for four independent reasons: they are actuated, so
energy is injected every step; they are dissipative through joint damping and
armature; they are contact-rich, so ground reaction forces break momentum
conservation and the vector field is non-smooth at touchdown; and even the
passive invariant would not be polynomial in the supplied coordinates, since
$\tfrac12 \dot q^\top M(q)\dot q$ is transcendental in the raw joint angles.
The Gymnasium observation additionally omits the root $x$-coordinate, so it is
not a state.

The correct behaviour of a discovery procedure on these environments is to
return nothing. We measure whether it does. A method that always finds
something is worse than useless in a safety pipeline, and false-positive
invariants are a documented failure mode in this literature. We additionally
run a synthetic stress test: inject a variable spuriously correlated on the
training trajectories but not causally constrained, and verify that
cross-trajectory validation and the rationality gate reject it.

These environments are retained as controls and not as targets, and no result in
this paper extends to them.

\subsection{The fault catalogue}
\label{sec:faults}

Section~VII of the main paper gives the reason for a fixed list. The categories
are chosen so that each stresses a different part of the
machinery. A \emph{parameter} fault moves the ideal while leaving it an ideal of
the same shape, so the recovered coefficients should name the offending
constant. A \emph{structural} fault removes a term from the equations of motion,
so the system stops conserving what it should and no parameter assignment
explains it. A \emph{numerical} fault changes the integrator or step size,
leaving the exact invariants untouched as algebra and moving only their drift.
An \emph{observation} fault breaks a bookkeeping identity rather than a physical
law.

\begin{table}[t]
\caption{The fifteen faults of Section~VII-A of the main paper, plus two healthy controls.
Every one is a mistake a real simulator or robot description has plausibly
carried. The \texttt{nips} Coriolis entry is not hypothetical but a variant
Gymnasium ships, so it is a structural fault with known ground truth that
no one wrote for this evaluation.}
\label{tab:faults}
\begin{center}
\footnotesize
\begin{tabular}{@{}p{0.28\linewidth}@{\hspace{5pt}}p{0.12\linewidth}@{\hspace{5pt}}
p{0.32\linewidth}@{\hspace{5pt}}>{\raggedright\arraybackslash}p{0.15\linewidth}@{}}
fault & category & description & must break \\
\hline
\texttt{acrobot\_m2\_1.3}   & parameter & second link mass $1.3$ not $1.0$
  & energy \\
\texttt{acrobot\_l1\_1.2}   & parameter & first link length $1.2$ not $1.0$
  & energy \\
\texttt{acrobot\_lc2\_0.6}  & parameter & second CoM at $0.6$ not $0.5$
  & energy \\
\texttt{acrobot\_I2\_1.5}   & parameter & second inertia $1.5$ not $1.0$
  & energy \\
\texttt{acrobot\_g\_9.81}   & parameter & gravity $9.81$ not $9.8$ & energy \\
\texttt{reacher\_l2\_0.13}  & parameter & $\ell_2 = 0.13$ not $0.11$
  & fk-$x$, fk-$y$ \\
\texttt{reacher\_l1\_0.09}  & parameter & $\ell_1 = 0.09$ not $0.10$
  & fk-$x$, fk-$y$ \\
\hline
\texttt{acrobot\_nips\_\allowbreak coriolis} & structural & the \texttt{nips} variant, a
  dropped Coriolis term & energy \\
\texttt{acrobot\_torque\_leak}   & structural & unmodelled constant torque
  $0.05$ & energy \\
\hline
\texttt{acrobot\_euler}     & numerical & explicit Euler substituted for RK4
  & energy \\
\texttt{acrobot\_dt\_0.2}   & numerical & the shipped $\Delta t = 0.2$, not
  the resolved $0.005$ & energy \\
\hline
\texttt{acrobot\_obs\_swap} & observation & $\sin\theta_1$, $\cos\theta_2$
  transposed & both unit-norm, energy \\
\texttt{acrobot\_obs\_\allowbreak unnormalised} & observation & $\cos\theta_1$ scaled by
  $1.02$ & unit-norm $\theta_1$, energy \\
\texttt{reacher\_dz\_offset} & observation & $10^{-3}$ $z$ leak into the
  fingertip vector & $d_z = 0$ \\
\texttt{reacher\_obs\_swap} & observation & $\cos\theta_2$, $\sin\theta_1$
  transposed & unit-norm, fk-$x$, fk-$y$ \\
\hline
\shortstack[l]{\texttt{none\_acrobot}\\\texttt{none\_reacher}} & control & correct dynamics,
  reseeded & nothing \\
\end{tabular}
\end{center}
\end{table}

Scrambling the observation breaks the energy as well as the unit-norm
constraints, because the energy is a polynomial \emph{in the observation}:
permute two of its arguments and the quantity computed is no longer the one that
is conserved. The table records it explicitly, because an under-specified
ground truth
would score the method against the wrong locus.

\section{Experimental Protocols and Full Results}
\label{app:experiments}

Every number below is produced by a script in the supplementary material; the mapping from result to script is in the README. The
enabling algebra results (Section~\ref{sec:e0}) need no training. The fault
catalogue carries the argument; the world-model and shaping results delimit it.

\subsection{Enabling results: deflation and the search}
\label{sec:e0}

\paragraph{Deflation.} Table~\ref{tab:deflation} reports the Acrobot degree-$3$
ablation on sampled trajectories. Normal-form reduction removes all $14$
algebraically trivial directions at every step size and tolerance, where
sequential orthogonal projection removes one direction \emph{in total} and not
one per generator. This holds for a simple reason: after $g_1$ has been projected out
the
subspace is $S \cap g_1^\perp$, which no longer contains $g_2$, and
$v \mapsto v - \langle v, g_2\rangle g_2$ is injective on any subspace holding
no vector parallel to $g_2$. Acrobot's two unit-norm generators both carry the
constant term $-1$, so their coefficient vectors have inner product $1$ and
this is exactly the degenerate case. Appendix~\ref{app:deflation} gives the
orthogonal-generator control, where the shortfall correctly does not appear.

The same table carries the negative reading developed in
Appendix~\ref{app:scaling}: at Gymnasium's shipped $\Delta t = 0.2$ the
dimension sequence $15 \to 1$ is indistinguishable from the one at
$\Delta t = 0.002$, yet $62\%$ of the survivor lies outside
$\mathrm{span}\{E,1\}$, so the count is not evidence of recovery and the
residual check is. That is the quantitative form of Remark~\ref{rem:tension} and
the reason the diagnostic uses the tolerance-free formulation of
Section~IV-B of the main paper.

\paragraph{Negative controls.} On Hopper ($n = 11$) and HalfCheetah ($n = 17$),
under a random policy over eight episodes and $4008$ samples, the degree-$1$ and
degree-$2$ nullspaces are empty at every tolerance from $10^{-10}$ to $10^{-6}$.
Hopper admits two directions at $10^{-4}$ and neither survives cross-episode
validation. The procedure returns nothing, which is correct. In a synthetic
stress test a variable $z = 2x$, correlated on the training trajectory but not
causally constrained, yields a relation whose held-out residual is
$1.2\times10^{-1}$ against $1.8\times10^{-17}$ for the genuine constraint
$x^2 + y^2 - 1$.

\subsection{Fault localisation and attribution: protocol and full results}

Section~VII-A of the main paper states the design: fifteen faults from
Table~\ref{tab:faults} with two reseeded healthy controls, in an unpaired and a
paired regime. Every setting each method runs under is in
Table~\ref{tab:e1protocol}.

\begin{table}[t]
\caption{Fault-catalogue protocol. Every setting is fixed across all methods and
all faults. The baselines subsample because their cost is superlinear in the
sample count, whereas the screen's is not. The MMD permutation budget sets that
test's smallest attainable $p$ at $1/10001$, two orders below $\alpha$, so the
budget cannot be what decides its verdict; the kernel is built once and permuted
by index, so that budget is affordable.}
\label{tab:e1protocol}
\begin{center}
\small
\begin{tabular}{@{}p{0.32\linewidth}@{\hspace{6pt}}p{0.60\linewidth}@{}}
setting & value \\
\hline
Acrobot trajectories & $8$ trajectories, $400$ steps, $\Delta t = 0.005$ \\
Reacher samples & $N = 4000$ poses, lengths scaled by $\ell_1$ \\
significance level $\alpha$ & $0.01$, Bonferroni corrected over $G_{\rm ref}$ \\
screen blocks & $32$ contiguous, $4$ per trajectory \\
screen trajectory units & $8$ per configuration, exact permutation null over
  $\binom{16}{8}$ splits \\
screen residual floor $\rho_{\min}$ & $10^{-8}$ \\
recovery spectral-gap threshold $\gamma$ & $0.1$ \\
attribution difference lag $k$ & $1$, $10$, $50$, $100$ \\
MMD & $n_{\rm sub} = 2000$, $10^4$ permutations, median-heuristic RBF \\
KS $+$ Bonferroni & $n_{\rm sub} = 2000$, asymptotic tail, per observation
  coordinate \\
MLP discriminator & $n_{\rm sub} = 3000$, two hidden layers of $64$, at most
  $300$ epochs \\
discriminator split & $50/20/30$ train/validation/test, early stopping on
  validation AUC, patience $20$ \\
discriminator decision rule & permutation $p < \alpha$ on held-out scores \\
paired difference & $32$ contiguous blocks, mean $|{\rm test} - {\rm ref}|$ per
  coordinate \\
paired difference null & the same statistic on a healthy paired pair,
  Bonferroni over coordinates \\
paired difference floor & $10^{-8}$, the same absolute floor the screen uses \\
\end{tabular}
\end{center}
\end{table}

The discriminator is scored by a test at $\alpha$ and not by a hand-picked
AUC cut. Because the AUC is computed on data the network never saw, the
exchangeability null is available without retraining: under the hypothesis that
the two samples come from one distribution the held-out labels are exchangeable
given the scores, so the permutation distribution of the AUC is the
Mann--Whitney null. That is the classifier two-sample test, and it puts every
baseline on one footing with the screen.

The split is three-way and training stops on the validation AUC, which is not
routine hygiene here but a correctness requirement. In the paired regime the two
configurations are near-duplicate rows carrying opposite labels. Trained for a fixed number
of epochs with no held-out stopping criterion, the network memorises the training
rows and generalises the inverse relation, returning a held-out AUC \emph{below}
chance: on two bit-identical configurations it scored $0.29$, which a one-sided test
reports as $p = 1$ and a two-sided one as overwhelming evidence that identical
data differs. Neither number measures the baseline's power. With early stopping,
the same two configurations score $0.506$, at chance, while a $30\%$ mass error still
scores $0.963$. The counts in Table~I of the main paper are lower than the memorising
form produced, and they are the ones that mean something.

A fourth baseline is included because the other three are handed less
information than the screen is. In the paired regime the configurations are logged from
identical initial conditions on a common grid, so row $i$ of one corresponds to
row $i$ of the other, and pooling them before a marginal two-sample test throws
that away. The paired difference test keeps it, and is built to mirror the
screen exactly: per-coordinate block-mean discrepancy in place of the
per-generator residual, the same $32$ blocks, the same one-sided rank test
against a healthy reference pair, the same Bonferroni family and the same
absolute floor. What separates it from the screen is the feature and nothing
else, which is the point.

\paragraph{Detection and localisation.} Section~VII-A of the main paper reads
Table~I of the main paper; two of its columns need the detail here. The paired difference tests detect every injected fault, and it is the screen's construction with the
per-generator residual replaced by a per-coordinate discrepancy, so what
separates them is the feature and nothing else. At $\sigma = 0$ its residual
ratio is infinite on every fault, a correct paired reference pair differing by
exactly zero, and under noise the ratios run from $1.23$ on Reacher's $d_z$ leak
to $782$ on a transposed Acrobot observation. Against that, it needs the
row-to-row correspondence, so it does not run in the unpaired regime at all,
which is the regime the release audit of Section~VII-D of the main paper faces.

\paragraph{The residual-feature ablation.} Section~VII-A of the main paper reports that the
baselines localise once they are handed the screen's features, the residual
vector whose $j$-th coordinate is the $j$-th reference generator's residual over
a block. The full counts: a two-sided Kolmogorov--Smirnov test over that vector,
Bonferroni corrected over the same family, localises $14/15$ against the
screen's $15/15$, and MMD's detection rises from $6/15$ on raw states to $14/15$
on residuals. The gain is not uniform, the discriminator moving only from $8/15$
to $9/15$, so what the residual features supply is a set of named univariate
channels a rank test can read one at a time, and a method pooling them into one
score gains little. The one fault the residual test misses is a
degenerate-column artefact: Reacher's $d_z$ residual is constant in both configurations,
so a two-sample test drops the column for having no distribution to compare even
though the configurations separate totally, whereas a one-sided test against zero would not.

\paragraph{Could a baseline localise after all?} The localisation column scores
agreement with a fault's \emph{constraint} locus, and no baseline has a
candidate to put there, though two come closer than the others. MMD and the
discriminator reduce the whole state vector to a single statistic, and there is
no map from that number back to a named object of any kind. KS and the paired
difference test compute a per-coordinate statistic that is reduced to a minimum
only at the end, so their readout could name an offending \emph{observation
coordinate}, and on the three observation faults either would do so cleanly,
naming the two transposed coordinates on \texttt{obs\_swap} and $d_z$ on
\texttt{r\_dz\_offset}.

What it cannot name is a constraint, for two separate reasons. On every
parameter and structural fault the error propagates into all coordinates within
a few steps, so the per-coordinate readout points everywhere and distinguishes
nothing. More fundamentally, a constraint is not a coordinate: Acrobot's energy
is a degree-$3$ polynomial in all six observation entries, and no subset of
coordinates answers ``which physical law failed''. A per-coordinate localisation
is a weaker output belonging in a column of its own.

\paragraph{Attribution.} On all seven single-parameter faults the correct parameter is
named, with worst relative error $1.0\times10^{-7}$ unpaired and
$1.7\times10^{-7}$ paired, over values spanning masses,
lengths, centres of mass, inertias, gravity and two link lengths in a separate
environment. Reacher's $\ell_2 = 0.13$ fault is returned as
$\ell_2/\ell_1 = 1.3000$ against a specification of $1.1$, which is the
diagnostic doing the thing that motivated the paper: it does not report that the
model is wrong, it reports which number is wrong and what it has become.

Equally important is where attribution declines. On the seven faults that carry
a parameterised generator and that no parameter assignment can explain, it names
no parameter and returns \emph{not explicable as a parameter fault} on every
one: the dropped Coriolis term of the \texttt{nips} variant, the unmodelled
constant torque, the substituted explicit-Euler integrator, the shipped
$\Delta t = 0.2$, the scaled $\cos\theta_1$, and the two transposed
observations. Reacher's $dz$ offset breaks only $d_z = 0$, which carries no
parameter, so level-2 does not apply to it and a refusal there would be scoring
the diagnostic on a question it was not asked.

Those seven are refused by two different mechanisms and the distinction
matters. Six fail at the spectral gap, no direction being admitted at all: the
dropped Coriolis term at $0.554$, the explicit-Euler integrator at $0.367$, the
shipped $\Delta t$ at $0.859$, the two transposed observations at $0.960$ and
$0.937$, and the unmodelled constant torque at $0.118$, all against
$\gamma = 0.1$ and all quoted unpaired. The scaled $\cos\theta_1$ is the
exception. It admits a direction easily, at a gap of $1.3\times10^{-7}$, and the
refusal comes from the \emph{fit} instead: the best single-parameter hypothesis
misses the absolute tolerance at $4.5\times10^{-3}$, so it does not explain the
recovered generator, which is evidence against the specification and not for it.
Pairing moves the torque leak across the same boundary, to a gap of $0.038$,
where it too is refused by the fit at a residual of $1.7\times10^{-2}$.

Separating those two failures is what makes the verdict correct, for the reason
Appendix~\ref{app:diagnostic} works through.

On both healthy controls at $\sigma = 0$ it returns \emph{consistent with
specification}, which is correct. Under noise the Acrobot control moves to
\emph{not explicable as a parameter fault}, which still names no parameter and
is the safe direction to fail in, though it is the less informative answer. A
procedure reporting the argmin unconditionally would instead name a fault on
correct software every time, because on a healthy system each single-fault
hypothesis fits well at its own specified value. It is the separation test of
Section~VI-B of the main paper, and not the fit alone, that carries the verdict.

\begin{table}[t]
\caption{Attribution under measurement noise: correct parameter named and
within $5\%$, out of seven single-parameter faults. The lag is the difference stride
of Section~IV-B of the main paper. Reacher's two faults are kinematic and use no
difference dictionary, which is why the count floors at two, not zero.}
\label{tab:attrnoise}
\begin{center}
\small
\begin{tabular}{r|cccc}
$\sigma$ & lag $1$ & lag $10$ & lag $50$ & lag $100$ \\
\hline
$0$        & $7$ & $7$ & $7$ & $7$ \\
$10^{-8}$  & $6$ & $6$ & $\mathbf{7}$ & $6$ \\
$10^{-7}$  & $2$ & $\mathbf{6}$ & $\mathbf{6}$ & $\mathbf{6}$ \\
$10^{-6}$  & $2$ & $2$ & $\mathbf{4}$ & $2$ \\
$10^{-5}$  & $2$ & $2$ & $2$ & $2$ \\
$10^{-4}$  & $2$ & $2$ & $2$ & $2$ \\
\end{tabular}
\end{center}
\end{table}

\paragraph{Detection floor against fault magnitude.} Every fault in
Table~\ref{tab:faults} sits at one magnitude, which answers whether a method
clears the bar set for it and nothing about where its floor lies.
Table~\ref{tab:floor} sweeps the magnitude instead: one constant is moved by a
relative amount from $10^{-4}$ to $0.3$ with everything else at specification,
and the smallest error each method still rejects at $\alpha$ is its floor on
that constant. The reference and test configurations are paired, so the difference between
them is the constant and not the draw.

\begin{table}[t]
\caption{Smallest relative parameter error still rejected at $\alpha = 0.01$,
swept over $\{10^{-4}, 3\times10^{-4}, \dots, 0.3\}$. ``none'' means no swept
magnitude was rejected. The screen is at the bottom of the sweep on all four
constants without noise; under noise it separates by generator kind, staying
near there on Reacher's vanishing constraints and rising by two to three orders
on Acrobot's conserved one.}
\label{tab:floor}
\begin{center}
\small
\begin{tabular}{llcccc}
noise & constant & MMD & KS & discriminator & \textbf{screen} \\
\hline
\multirow{4}{*}{$\sigma = 0$}
 & Acrobot $m_2$    & $0.3$ & $0.3$ & $0.3$ & $\mathbf{\le 10^{-4}}$ \\
 & Acrobot $g$      & $0.3$ & $0.3$ & $0.3$ & $\mathbf{\le 10^{-4}}$ \\
 & Acrobot $l_{c2}$ & $0.3$ & $0.3$ & $0.3$ & $\mathbf{\le 10^{-4}}$ \\
 & Reacher $\ell_2$ & none & none & none & $\mathbf{\le 10^{-4}}$ \\
\hline
\multirow{4}{*}{$\sigma = 10^{-3}$}
 & Acrobot $m_2$    & $0.3$ & $0.3$ & $0.3$ & $\mathbf{0.1}$ \\
 & Acrobot $g$      & $0.3$ & $0.3$ & $0.3$ & $\mathbf{0.03}$ \\
 & Acrobot $l_{c2}$ & $0.3$ & $0.3$ & $0.3$ & $\mathbf{0.1}$ \\
 & Reacher $\ell_2$ & none & none & none & $\mathbf{10^{-3}}$ \\
\end{tabular}
\end{center}
\end{table}

The screen's floor without noise is below the bottom of the sweep on every
constant, so the table bounds it rather than measuring it; a $10^{-4}$ relative
error in $m_2$ still produces a residual ratio of $5.5\times10^{5}$. On this
sweep the screen leads on every constant at both noise levels, which is a
stronger ordering than the fault catalogue shows, and the reason is that the
sweep is paired and one constant moves at a time. Under noise the interesting
structure is not between methods but between kinds of generator. A vanishing
constraint is evaluated pointwise and its residual rises with the fault while
the noise floor stays put, which is why Reacher holds at $10^{-3}$. A conserved
quantity is measured as a drift across a window, and observation noise enters
that drift directly, which is why Acrobot's floor rises by two to three orders.

\paragraph{Operating characteristics over $\alpha$.} Table~I of the main paper is one
draw at one significance level, so it says nothing about how the ordering
between methods behaves as the level moves. Table~\ref{tab:roc} sweeps
$\alpha$ over twelve decades on the same catalogue, scoring every method at the
trajectory unit so that each is read against an exact null. The screen goes
from nothing to everything at $\alpha = 10^{-3}$ and stays there, whereas the
three baselines are flat across the whole range: the discriminator sits at
$0.53$ from $10^{-12}$ to $0.05$ without moving once, MMD climbs from $0.33$ to
$0.53$ and stops, and Kolmogorov--Smirnov reaches $0.40$. A curve that does not
respond to its own threshold is not being held back by the level it is scored
at, so the gap between the screen and the baselines is not a matter of where
$\alpha$ was put.

Two limits bound how far down the table can be read, and both are properties of
the nulls and not of the systems. The exact permutation null over eight
trajectories per configuration cannot return a $p$-value below
$\binom{16}{8}^{-1} = 7.8\times10^{-5}$, which the three-generator correction
takes to $2.3\times10^{-4}$; MMD's permutation $p$ floors at $1/(B+1)$ with
$B = 10^{4}$. Rows beneath a method's floor are marked, and a zero there is
arithmetic and carries no information about power. The false-positive rate is
$0.00$ for every method at every level in the table, but it rests on the two
healthy controls, so the measurement carries a Wilson upper bound of $0.66$ and
Fig.~3 of the main paper is where the rate is actually estimated.

\begin{table}[t]
\caption{True-positive rate over the fifteen catalogue faults as the
significance level sweeps twelve decades, scored at the trajectory unit. An
asterisk marks a level below the method's $p$-value floor, where a rate of
zero is arithmetic. The false-positive rate is $0.00$ over the two healthy
controls for every method and every level shown, which two controls bound only
at $[0, 0.66]$; the rate itself is measured in
Fig.~3 of the main paper.}
\label{tab:roc}
\begin{center}
\small
\begin{tabular}{llcccc}
noise & $\alpha$ & MMD & KS & discriminator & \textbf{screen} \\
\hline
\multirow{6}{*}{$\sigma = 0$}
 & $10^{-12}$ & $0.00^{*}$ & $0.13$ & $0.53$ & $0.00^{*}$ \\
 & $10^{-6}$  & $0.00^{*}$ & $0.20$ & $0.53$ & $0.00^{*}$ \\
 & $10^{-4}$  & $0.33$ & $0.20$ & $0.53$ & $0.00^{*}$ \\
 & $10^{-3}$  & $0.33$ & $0.20$ & $0.53$ & $\mathbf{1.00}$ \\
 & $10^{-2}$  & $0.40$ & $0.40$ & $0.53$ & $\mathbf{1.00}$ \\
 & $0.05$     & $0.53$ & $0.40$ & $0.53$ & $\mathbf{1.00}$ \\
\hline
\multirow{6}{*}{$\sigma = 10^{-3}$}
 & $10^{-12}$ & $0.00^{*}$ & $0.20$ & $0.53$ & $0.00^{*}$ \\
 & $10^{-6}$  & $0.00^{*}$ & $0.27$ & $0.53$ & $0.00^{*}$ \\
 & $10^{-4}$  & $0.33$ & $0.27$ & $0.53$ & $0.00^{*}$ \\
 & $10^{-3}$  & $0.33$ & $0.27$ & $0.53$ & $\mathbf{0.80}$ \\
 & $10^{-2}$  & $0.40$ & $0.47$ & $0.53$ & $\mathbf{0.80}$ \\
 & $0.05$     & $0.53$ & $0.47$ & $0.53$ & $\mathbf{0.80}$ \\
\end{tabular}
\end{center}
\end{table}

\paragraph{The noise ceiling.} Table~\ref{tab:attrnoise} bounds attribution, and
Section~VII-A of the main paper draws the division of labour that follows from it. The lag
of Section~IV-B of the main paper is what provides the decade: at $\sigma = 10^{-7}$ it
takes the count from $2/7$ to $6/7$, and at $\sigma = 10^{-6}$ from $2/7$ to
$4/7$. The gain saturates once secular drift over the window becomes comparable
to the noise, so a longer window provides nothing further.

\subsection{Algebraic unit tests for learned world models}

Eighty-four MLP dynamics models on Acrobot, varying the consistency weight
$\lambda$ over $\{0,\allowbreak 10^{-3},\allowbreak 10^{-2},\allowbreak 10^{-1},\allowbreak 1,\allowbreak 10,\allowbreak 100\}$ together with
hidden width, trajectory budget and seed, so that the comparison between
predictors is measured over a genuine spread of model quality rather than
asserted from two points. Fidelity is scored by \emph{divergence horizon}, the
first step at which rollout RMSE against ground truth exceeds a threshold;
terminal error saturates on a system this sensitive and stops separating good
models from bad ones. The threshold is swept from $0.05$ to $1.0$ because the
conclusion rests on it, and $0.25$ is quoted where a single value is needed.

Section~VII-B of the main paper gives the correlations and the pools they are computed
over. Two details of the design belong here. The twelve unregularised models are
the estimate and the full sweep bounds the gap from above, since the other
seventy-two were trained with the algebraic score in the loss, which inflates
one predictor and deflates the other. And what survives is weaker than the
hypothesis but not nothing: the algebraic score carries real signal while
consulting \emph{no reference at all}, being computed from the model's own
imagined rollouts, which is the situation on any system whose long-horizon
ground truth is the missing quantity.

\begin{table}[t]
\caption{Consistency weight against what it moves, paired within architecture,
data budget and seed, $12$ pairs per row against $\lambda = 0$. Ratios are
geometric means. The interval is a percentile bootstrap on the mean paired
difference in divergence horizon.}
\label{tab:lambda}
\begin{center}
\small
\begin{tabular}{lcccc}
$\lambda$ & val MSE & algebraic residual & horizon & $95\%$ interval \\
\hline
$10^{-3}$ & $1.000$ & $0.997$ & $-0.1$ & $[-0.5, +0.2]$ \\
$10^{-2}$ & $1.008$ & $1.000$ & $-1.3$ & $[-2.5, -0.1]$ \\
$10^{-1}$ & $1.036$ & $0.993$ & $-2.1$ & $[-6.0, +1.7]$ \\
$1$       & $1.574$ & $0.834$ & $-4.0$ & $[-8.2, +0.5]$ \\
$10$      & $7.540$ & $0.874$ & $-20.0$ & $[-24.4, -15.9]$ \\
$100$     & $64.16$ & $0.787$ & $-36.8$ & $[-44.7, -29.0]$ \\
\end{tabular}
\end{center}
\end{table}

Table~\ref{tab:lambda} is the paired comparison, and one feature of it decides
how far the weight can be pushed. Even at $\lambda = 100$, where one-step error
is $64$ times its unregularised value, the algebraic residual has only fallen to
$0.79$, because the penalty acts on one-step predictions while the residual is
measured over a hundred-step rollout, and no weight in this range crushes the
second without destroying the first.

Sweeping the divergence threshold from $0.05$ to $1.0$ separates the table's two
findings instead of confirming both. The collapse survives the sweep intact,
running from $-8.5$ to $-20.0$ steps at $\lambda = 10$ and from $-12.9$ to
$-44.1$ at $\lambda = 100$. The three smallest weights change sign across the
range: $\lambda = 10^{-1}$ runs $+0.0, -0.5, -2.1, -0.5, -0.2$,
$\lambda = 10^{-2}$ runs $+1.1, -0.7, -1.3, +0.6, +0.5$ and
$\lambda = 10^{-3}$ runs $+0.1, -0.1, -0.1, +0.1, +0.1$. The single interval
excluding zero is therefore a property of the threshold it was read at, and what
the sweep leaves standing is the collapse.

\subsection{Shaping: protocol, controls and full curves}

Acrobot swing-up with PPO \citep{schulman2017ppo}, five seeds, $120{,}000$
environment steps, reporting
\emph{unshaped} environment return throughout; reporting shaped return would
make any bonus an improvement by definition. The potential is
$\Phi(s) = -|E(s) - E^\star|$ with $E$ \emph{discovered}, not assumed:
Section~IV-B of the main paper is run on finely re-integrated passive trajectories
and the recovered polynomial defines the potential, which is then deployed on
Acrobot-v1 as distributed. That detour is forced by Remark~\ref{rem:tension},
since at the shipped $\Delta t = 0.2$ the energy is not recoverable, and it is
a real constraint on the pipeline.

Four kinds of condition, each ruling out a specific alternative explanation.
\textbf{none} is PPO on the environment reward. \textbf{discovered} uses the
recovered energy. \textbf{faulty} uses a potential recovered from a system whose
second link mass is wrong by $10$, $30$, $100$ or $300$ percent, which asks
whether the diagnostic provides anything downstream and, across the sweep, how wrong
the physics may be before it stops. \textbf{random} uses a degree-$3$ polynomial
with no physical meaning, rescaled to the same range, which asks whether any
dense signal would do. \textbf{rnd} is Random Network Distillation
\citep{burda2019rnd}, a bonus that is not potential-based and carries no
policy-invariance guarantee.

Discovery costs a fraction of a second and returns the energy to a relative
coefficient error of $2.8\times10^{-8}$ against the analytic form, so the
potential is recovered, not supplied.

\paragraph{The gauge, and why the faulty configuration must fix its own.} Recovery returns
a direction, so a recovered potential carries an arbitrary scale and offset and
neither is deployable: $\Phi(s) = -|E(s) - E^\star|$ is meaningless until $E$
and
$E^\star$ are in the same units. Each condition therefore fixes its gauge by
least squares over $4000$ random states, and the reference it regresses against
is the analytic energy of \emph{its own} model. For the faulty configurations that is the
faulty model's energy, and the target $E^\star$ is that model's upright
pose:
$21.07$, $24.01$, $34.30$ and $63.70$ across the sweep, against the correct
$19.60$. A practitioner deploying from a model they have not checked has exactly
this and nothing more, so the parameter error is carried in the coefficients, in
the scale, in the offset and in the target at once.

The \textbf{random} configuration needs a different gauge, and which one decides whether
the control is informative. Regressing the random polynomial onto the energy, as
the faulty configurations regress onto their own, is the wrong operation here: the two are
uncorrelated, so the fitted slope is driven by a correlation that does not exist
($R^2 = 2.3\times10^{-5}$) and collapses towards zero, returning a potential
that is constant to within two orders of magnitude and contributes no shaping
signal at all. Matching moments fixes the same two degrees of freedom and
delivers a scale, but it leaves open which sample the moments are taken over and
which moment is matched, and both choices move the deployed potential.

The sample has to be the states a run visits. Acrobot-v1 admits velocities out
to $4\pi$ and $9\pi$, a degree-$3$ polynomial extrapolates as a cube, and two
potentials matched over a box of independently drawn angles and velocities
separate off it: the same polynomials then deploy at five times the energy's
range. Moments are therefore taken over $20{,}000$ states from a uniform-random
policy on the shipped environment, which is the region a run starts in.

The moment has to be chosen, because no scale and offset match both. Over those
states the energy has spread $5.75$ and hands the agent a per-step
$\gamma\Phi(s') - \Phi(s)$ of standard deviation $0.748$; a random polynomial
matched to the first has an increment of about $4.0$, and one matched to the
second has spread about $1.07$. That is not a defect of the gauge, it is what
being conserved means: the energy moves little between consecutive states while
ranging widely over the state space, and no arbitrary polynomial does both.
Since the increment is what is added to a reward of $-1$ per step, and the range
is what the potential is nominally matched on, the control is run under both
matchings and Table~\ref{tab:shaping} reports both quantities for every deployed
potential.

\begin{table}[t]
\caption{Acrobot swing-up, PPO, five seeds, $120{,}000$ steps, unshaped
environment return, higher better. AUC is the mean return over a common step
grid. \emph{Spread} is the standard deviation of the deployed potential over
states a random policy visits, and \emph{step} the standard deviation of the
$\gamma\Phi(s') - \Phi(s)$ it hands the agent there, against an environment
reward of $-1$ per step. $\Delta$ is the AUC difference from the discovered
potential with a percentile bootstrap interval. The $p$ column flags detection
against \emph{none} only, uncorrected across conditions; five seeds against five
bottom out at $0.0079$ where ties do not force the asymptotic form. Random rows
give the mean and range of ten draw means, not a pooled seed-level interval or
$p$-value. RND runs at a single un-swept bonus coefficient.
Every faulty potential is recovered from a faulty system and gauged entirely
within it.}
\label{tab:shaping}

\footnotesize
\setlength{\tabcolsep}{1.8pt}

\begin{tabularx}{\textwidth}{@{}>{\raggedright\arraybackslash}X
  r r r r >{\raggedright\arraybackslash}X r@{}}
condition & AUC & final return & spread & step & $\Delta$ vs.\ discovered & $p$ \\
\hline
none (PPO)
  & $-147.8 \pm 28.4$
  & $-84.4 \pm 4.1$
  & & & $-56.5$ $[-81.6, -31.9]$
  & \\

rand. matched range
  & $-195.3$ $[-443.6, -126.7]$
  & $-132.7$ $[-416.5, -85.7]$
  & $5.75$ & $4.05$
  & $-104.0$ $[-352.3, -35.4]$
  & \\

rand. matched step
  & $-162.2$ $[-214.0, -135.1]$
  & $-99.1$ $[-169.6, -83.5]$
  & $1.06$ & $0.75$
  & $-70.9$ $[-122.7, -43.8]$
  & \\

RND \citep{burda2019rnd}
  & $-118.1 \pm 8.2$
  & $-85.8 \pm 3.1$
  & & &
  $-26.7$ $[-33.7, -19.0]$
  & $0.22$ \\

\textbf{discovered}
  & $\mathbf{-91.3 \pm 2.4}$
  & $-84.5 \pm 3.1$
  & $5.75$ & $0.75$
  & & $0.0079$ \\

faulty, $m_2$ $+10\%$
  & $-90.8 \pm 1.3$
  & $-83.7 \pm 2.2$
  & $6.07$ & $0.79$
  & $+0.5$ $[-1.5, +3.1]$
  & $0.0079$ \\

faulty, $m_2$ $+30\%$
  & $\mathbf{-88.6 \pm 1.6}$
  & $-85.3 \pm 3.1$
  & $6.71$ & $0.90$
  & $+2.7$ $[+0.4, +5.3]$
  & $0.0079$ \\

faulty, $m_2$ $+100\%$
  & $-90.8 \pm 1.9$
  & $-89.1 \pm 5.4$
  & $9.00$ & $1.49$
  & $+0.5$ $[-2.0, +3.2]$
  & $0.0079$ \\

faulty, $m_2$ $+300\%$
  & $-100.5 \pm 6.0$
  & $-88.1 \pm 4.0$
  & $15.68$ & $3.54$
  & $-9.2$ $[-15.2, -4.0]$
  & $0.0079$ \\
\end{tabularx}
\end{table}

Section~VII-C of the main paper reads Table~\ref{tab:shaping} and is not repeated here. One
number the main body omits belongs with the sweep: the $300$ percent condition
is separable from the correct potential at $p = 0.016$, while remaining better
than no shaping at $p = 0.0079$, so its degradation is resolved rather than
merely apparent.

RND sits between the two. Its AUC of $-118.1$ is not separable from no shaping
over five seeds ($p = 0.22$), but a bootstrap interval on the difference in
means, $[+4.7, +55.3]$, excludes zero, against the $+56.5$ the discovered
potential delivers. Five seeds cannot support a claim that the two are
equivalent, so what the comparison supports is an ordering: the discovered
potential's gain over no shaping is about twice RND's, and the interval is too
wide to say more.
\subsection{The release audit: matrix, protocol and findings}

Section~VII-D of the main paper gives the design and the findings. This section carries the
release matrix, the per-step amplification trace, the controls and the
source-level diagnosis of the Reacher discrepancy.

\paragraph{The release matrix, and how to obtain it.} Eleven environments are
audited over two release lists, and every release named below is a published
version on the Python Package Index, installed at that exact version into a
virtualenv of its own on CPython 3.10.

The classic-control list holds eleven releases: \texttt{gym} 0.23.1, 0.25.2 and
0.26.2, followed by \texttt{gymnasium} 0.26.3, 0.27.1, 0.28.1, 0.29.1, 1.0.0,
1.1.1, 1.2.3 and 1.3.0. Pairing each with each gives $\binom{11}{2} = 55$ pairs
for every one of Acrobot-v1, Pendulum-v1, CartPole-v1, MountainCar-v0 and
MountainCarContinuous-v0, which is $275$ pairs. The MuJoCo list holds six
binding configurations, written as (\texttt{gymnasium}, \texttt{mujoco}):
$(0.29.1, 2.3.7)$, $(1.0.0, 3.1.6)$, $(1.2.3, 3.2.7)$, $(1.3.0, 3.1.6)$,
$(1.3.0, 3.2.7)$ and $(1.3.0, 3.10.0)$. The last three hold \texttt{gymnasium}
fixed and move only the binding, so that a difference could be attributed to the
physics library and not to the wrapper. That list covers Reacher,
InvertedPendulum and InvertedDoublePendulum in versions v4 and v5. The v4
environments are paired over all six configurations, $\binom{6}{2} = 15$ pairs
each, while the v5 environments are paired over only the five configurations
from \texttt{gymnasium} 1.0.0 that ship them, $\binom{5}{2} = 10$ pairs each.
That is $75$ MuJoCo pairs, and $350$ in total.

\texttt{gym} 0.21.0 is absent because it pins an \texttt{opencv-python}
specifier that modern resolvers reject, and 0.23.1 covers the same pre-0.26 API
era. NumPy is pinned per release, not globally, since releases before
\texttt{gym} 0.26 use \texttt{np.bool8} and fail on NumPy 1.24 and later. The
resolved NumPy version is written into every trajectory dump, so that a change
attributable to NumPy can be separated from a change in the dynamics.

Counting v4 and v5 separately, as the audit does, six of the eleven carry an
exact polynomial reference set: Acrobot (two unit-norm identities and energy),
Pendulum (unit-norm and energy), Reacher-v4 and Reacher-v5 (forward kinematics),
and InvertedDoublePendulum-v4 and v5 (two unit-norm identities each). The
remaining five carry none. CartPole's action set has no zero element, so the
cart is driven every step; both MountainCar variants sit in a cosine potential
that is transcendental in the position coordinate; and InvertedPendulum reports
its hinge angle raw rather than as a $(\sin,\cos)$ pair, so it carries no
embedding identity either. Those five are audited by the exact stage alone,
which is the stage that produces the null; the screen exists to localise a
separation, and where there is none there is nothing to localise.

\paragraph{The result is a null, across everything tested.} All $275$
classic-control pairs are bitwise identical, at a maximum separation of exactly
zero. Among the $75$ MuJoCo pairs, $26$ separate at float noise below
$10^{-13}$, the largest off the chaotic environment being $9.1\times10^{-15}$,
consistent with a recompiled physics library reordering floating-point
operations without changing the model.

\paragraph{Chaos and the reading of an exact comparison.} Section~VII-D of the main paper
gives the finding; Fig.~4 of the main paper is the step-by-step trace
behind it.

On the thirteen InvertedDoublePendulum pairs the separation runs from
$2.2\times10^{-16}$ at step $0$ through $4.8\times10^{-9}$ at step $100$ to
$6.4\times10^{-2}$ at step $400$, smooth exponential growth at one decade per
$32.4$ steps on the first pair and per $41.6$ on the second, fitted between
steps $50$ and $400$ where the separation is off the last-place staircase and
below the clip. That is a Lyapunov exponent acting on a rounding difference and
not the discontinuity a changed constant would produce. Reacher, traced over the identical release pairs and
horizon, sits flat at $5.6\times10^{-17}$ from step $0$ to step $499$ on one
pair and at exactly zero on the other, isolating the cause to the dynamics and
not to the binding or the wrapper. Loosening the threshold to compensate would
give up the certainty that motivated putting an exact comparison first, so the
remedy is a shorter comparison window and not a larger tolerance.

\paragraph{The controls run on the shipped class.} A null means nothing unless
the same pipeline separates a difference known to be present, so the controls
perturb Gymnasium's \texttt{AcrobotEnv} and not our reimplementation. At the
shipped $\Delta t = 0.2$, the exact stage separates both controls on
byte-identical inputs: selecting \texttt{book\_or\_nips=nips}, a variant
Gymnasium ships that drops a Coriolis term, moves the trajectory by a maximum
$|\Delta| = 10.0$, and setting \texttt{LINK\_MASS\_2} to $1.3$ moves it by
$9.7$. That is the exact stage producing the audit's null doing its work on
code we did not write.

Localisation at the shipped step size is not demonstrated on these controls,
and cannot be. The screen runs at the trajectory unit, one residual per whole
trajectory with an exact permutation null over the $\binom{16}{8}$ splits of
eight trajectories per configuration, which is the unit the stated $\alpha$ is
supported at (Section~VIII of the main paper). Under that test the two controls
give energy residual ratios of $1.51$ at $p = 0.34$ and $3.32$ at $p = 0.017$,
neither below $\alpha = 0.01$, and neither localises.

At the block unit the same two controls do localise, with energy residual
ratios of $4.21$ at $p = 1.7\times10^{-6}$ and $11.56$ at
$p \le 9.8\times10^{-12}$, both localising to \texttt{energy}. That unit is
not the one reported in Table~II of the main paper, because the block-level test
does not hold its nominal level (Section~VIII of the main paper): its $32$ short
windows per configuration share an initial condition and are not the
independent replicates the rank test counts them as, and the measured
block-level false-alarm rate reaches $10.2\%$ against a nominal $1\%$. The
values in Table~II of the main paper are trajectory-unit, which is the unit the
stated $\alpha$ is supported at.

\paragraph{What limits the screen here is the simulator, not the screen.}
Section~VII-D of the main paper states the finding. Holding the simulated horizon fixed and
reducing the step moves the drift floor and the detection threshold together, as
Table~II of the main paper shows.

This is Remark~\ref{rem:tension} measured on software in the field and not on
our own integrator, the two agreeing on the mechanism while differing in
absolute size because the audit accumulates drift over a $100\,$s horizon
against the remark's $40\,$s. What the audit adds is the consequence for
diagnosis: the remark shows the window for \emph{discovery} has closed by the
shipped step size, and Table~II of the main paper shows \emph{screening} degrades
with it, from localising a $0.1\%$ error at $\Delta t = 0.05$ to localising
nothing up to a $30\%$ error at the shipped $\Delta t = 0.2$.

\paragraph{Two floors that bound level-2 attribution in practice.} The audit
turns the noise ceiling of Section~VII-A of the main paper from an abstract quantity into
two concrete numbers. Acrobot and Pendulum return float32 observations, which
puts a floor of about $1.7\times10^{-8}$ on the unit-norm identities no matter
how exact the underlying dynamics are; the MuJoCo environments return float64
and reach $3.6\times10^{-17}$ on the same identities. Since attribution degrades
above a relative noise of $10^{-7}$, the observation dtype alone places
attribution on shipped classic-control environments at the edge of its ceiling.
Reading \texttt{env.unwrapped.state} instead of the observation removes that
floor entirely, and we recommend it.

\paragraph{The audit found something we did not put there.} Section~VII-D of the main paper
gives the Reacher finding and its source-level cause. Three measurements support
it. Evaluated after \texttt{mj\_forward} the identity holds at
$2.1\times10^{-17}$, so \texttt{reacher.xml} and the idealised two-link identity
agree exactly and the geometry is not at fault. The residual appears only once
the joints move during a step, at a median of $6.2\times10^{-9}$ and a maximum
of $2.3\times10^{-8}$. And \texttt{mj\_step} is called by \texttt{MujocoEnv.\_\allowbreak step\_\allowbreak mujoco\_\allowbreak simulation}
 followed by \texttt{mj\_rnePostConstraint}, never
\texttt{mj\_forward}. This is why \texttt{data.xipos} still describes an
internal integrator stage while \texttt{qpos} has been integrated.

The consequence for this instrument is a floor of roughly $10^{-8}$ on any exact
diagnostic applied to shipped MuJoCo observations, one order inside the
attribution ceiling of Table~\ref{tab:attrnoise}.

One sampling detail bounds the measurement. Sampling $\theta_2$ over
$[-\pi, \pi]$ places $(\pi-3)/\pi \approx 4.5\%$ of draws outside
\texttt{joint1}'s $[-3, 3]$ limit, where MuJoCo's limit constraint legitimately
moves \texttt{qpos} during the step and inflates the maximum residual to
$2.5\times10^{-4}$. That is the constraint working correctly, not the
inconsistency described above, and every number quoted here is measured inside
the limit.

\section{Verified Environment and Algebraic Properties}
\label{app:verified}

All statements below are outputs of the verification driver in the accompanying
repository (Python 3.14, SymPy $1.14$, NumPy $2.5$), which runs $143$ checks and
exits nonzero if any claim in this appendix has drifted from the text. Every
numerical check compares against the figure printed here to a relative
tolerance of $5\%$, rather than against an order-of-magnitude bound, so a value
that drifts inside its own order of magnitude still fails. The repository README
maps each claim to the script that establishes it. None is an experimental
result.

Six of the checks guard the diagnostic itself, not the environments. The
Acrobot energy exists twice in the codebase, once numerically and once
symbolically so that attribution can differentiate it with respect to physical
parameters, and a duplicated definition is where two things drift apart; the
check asserts they agree to $2.8\times10^{-14}$ over random states. The same is
asserted for Reacher's forward-kinematics template, which vanishes to
$8.9\times10^{-16}$ on generated poses. A third runs the full level-2 procedure
on a healthy system and asserts that the verdict is not \emph{parameter fault},
since a diagnostic that accuses correct software is worse than no diagnostic.
The remaining three assert that the trivial filtered piece has dimension $14$,
that the undeflated least-varying direction lies entirely inside it with zero
alignment to the energy, and that the deflated one aligns with the energy to
$1.0000$.

A further block of eight guards behaviour that a results table cannot see,
because each concerns a case the reported experiments do not visit. The two
conditions on a level-2 verdict are asserted separately, since a recovered
generator that is well separated but that no hypothesis fits must return
\emph{not explicable as a parameter fault} and not \emph{consistent with
specification}; the check builds exactly that case by perturbing an $m_2 = 1.3$
energy along a monomial outside the template's support. The residual floor is
asserted to be absolute by screening a $d_z$ leak of $10^{-30}$ and requiring
that it is \emph{not} flagged while a leak of $10^{-3}$ is, which is the
distinction a scale-relative floor cannot draw on a single-term generator.
The numerical nullspace of a $3 \times 10$ design matrix is asserted to have all
seven of its dimensions, and to satisfy $\Phi v = 0$ on each. And the random
control of Section~VII-C of the main paper is asserted to have the same spread over the
state space as the energy, to one percent, so that ``matched in scale'' remains
a measured property of the potential actually deployed.

A last group covers quantities the results tables have no column for. The
shaping section's gauge is one: its recovered energy, its agreement with the
analytic form and the five targets $E^\star$ come out of the discovery step and
not out of the training run, so none of them is a column of the returns table
and each is recomputed and compared here. The chaotic growth rate is another.
Both the rate and the number of decades climbed are quoted in
Section~VII-D of the main paper and in Fig.~4 of the main paper, and both are written
into the trace the figure is drawn from and read back, since a figure computed
only for a terminal is one that nothing can contradict. Table~\ref{tab:roc} is
compared cell by cell against the sweep that produced it, and its censoring
marks are derived from each method's $p$-value floor instead of trusted,
because a mark on the wrong row turns a measurement into arithmetic or the
reverse.

\paragraph{A model detail that determines everything else.} Gymnasium's Acrobot
offers two dynamics variants. The default, \texttt{book}, carries the term
$-m_2 l_1 l_{c2}\dot\theta_1^2\sin\theta_2$ in the second equation of motion;
the \texttt{nips} variant omits it. Only the former conserves the energy
\eqref{eq:acrobotE}. Under the latter, passive energy drift plateaus near
$1.7\times10^{-4}$ independently of $\Delta t$, which presents as an integrator
failure but is a model inconsistency, and every drift and balance number below
would be meaningless under it. The discrepancy is invisible until one measures
convergence order rather than drift magnitude, which is why the variant is
pinned explicitly rather than left at whatever the installed release defaults
to.

\subsection{Acrobot energy as a polynomial in the observation}
\label{app:acrobot}
Equation~\eqref{eq:acrobotE} is derived symbolically from the mass matrix and
potential, not asserted, and the derivation recovers exactly the eight
terms printed there. Over $2000$ uniformly random states the maximum discrepancy
between \eqref{eq:acrobotE} evaluated on the observation vector and the
mechanical energy $T + V$ evaluated in $(\theta,\omega)$ coordinates is
$1.14\times10^{-13}$, i.e.\ machine precision. The coefficient denominators are
$\{2,4,8,10\}$, so the largest is $10 \le Q_{\max} = 16$ and the invariant
would remain representable under the rounding scheme of
Section~\ref{sec:units} without rescaling. The ceiling of $16$ is not derived
from a stated precision or sample-size criterion; any value from $10$,
Acrobot's own maximum, up to $99$, one below Reacher's raw-unit denominator of
$100$ (Appendix~\ref{app:reacher}), separates the two cases identically.

\subsection{Passive drift under RK4}
\label{app:drift}
Relative energy drift $|\Delta E| / |E_0|$ over a $40\,$s passive rollout from
$(\theta,\omega) = (0.1, -0.2, 0.3, -0.15)$: $6.51\times10^{-4}$ at
$\Delta t = 0.2$
(Gymnasium's default); $6.86\times10^{-7}$ at $\Delta t = 0.05$;
$2.20\times10^{-10}$ at $\Delta t = 0.01$. The implied convergence order is
$4.95$, consistent with RK4. This is the drift floor of
Remark~\ref{rem:tension}, and it is the quantity that bounds the resolution of
any diagnostic built on a conserved quantity.

\subsection{Power balance under torque}
\label{app:balance}
With $\tau = 1$, the mean residual of
$E(s') - E(s) - \Delta t\,\tau\,\omega_2$ over $200$ steps is
$2.48\times10^{-3}$ at $\Delta t = 0.05$, $1.06\times10^{-4}$ at
$\Delta t = 0.01$ and $6.12\times10^{-6}$ at $\Delta t = 0.002$; the ratio
residual$/\Delta t^2$ is $0.99$, $1.06$ and $1.53$ respectively, a spread of
$1.5\times$ over a $25\times$ range of step sizes, which confirms a first-order
identity with $O(\Delta t^2)$ discretisation error.

\subsection{The twisted cubic}
\label{app:tc}
For $I = \langle y - x^2,\ z - x^3\rangle$ the reduced Gr\"obner basis under
grevlex with $x > y > z$ is $\{x^2 - y,\ xy - z,\ y^2 - xz\}$. The ideal
generated by the first two elements equals the ideal generated by all three, so
the third is redundant as a generator: three basis elements for a two-generator
ideal. This witnesses Remark~\ref{rem:notminimal}.

\subsection{Deflation on Acrobot}
\label{app:deflation}
With $n = 6$ and $d = 3$ the dictionary has $M = \binom{9}{3} = 84$ monomials.
The degree-$\le 3$ filtered piece of $\langle g_1, g_2\rangle$ has dimension
$14$, namely the products $m \cdot g_j$ with $\deg m \le 1$ and
$j \in \{1,2\}$; adjoining the energy direction gives $15$. Reducing modulo the
reduced Gr\"obner basis of $\langle g_1, g_2\rangle$ sends all fourteen to zero
and leaves the energy direction non-zero, giving a deflated nullspace of
dimension $1$. That much is exact algebra and holds at any tolerance;
Table~\ref{tab:deflation} reports what happens on sampled trajectories, where
it does not.

\paragraph{Sequential projection saturates.} On the same system, projecting
$g_1$ out of the $14$-dimensional component and then $g_2$ out of the result
removes one direction in total, leaving $13$. Both generators carry the
constant term $-1$, so their coefficient vectors have inner product $1$, and
after the first projection $g_2$ no longer lies in the subspace. The control
case confirms that this is the cause, not a numerical artefact: for the
orthogonal generators $h_1 = xy$ and $h_2 = zw$ in $n = 4$ variables, whose
coefficient vectors have inner product $0$, the same sequential projection
removes one direction per generator as the informal argument predicts, taking
the degree-$\le 3$ component from $10$ to $8$. Normal-form reduction removes all
$14$ and all $10$ respectively, in both cases.

\subsection{Reacher forward kinematics and the units failure}
\label{app:reacher}
Over $5000$ random poses the maximum residual of \eqref{eq:reacherfk} is
$7.29\times10^{-17}$. In metres the coefficients are $1/10$ and $11/100$, with
denominators $10$ and $100$; at $Q_{\max}=16$ the second is not representable
and the invariant is not recoverable under that ceiling, which is why the
scheme of Section~\ref{sec:units} nondimensionalises rather than working in
raw units. As in Appendix~\ref{app:acrobot}, $16$ is one admissible choice
among several rather than a derived bound. Rescaling lengths by $\ell_1$ gives
$1$ and $11/10$, with denominators $1$ and $10$, both admissible; the division
is carried out in floating point and the result recovered by an uncapped
rational reconstruction rather than by exact arithmetic, unambiguous here but
not itself checked against $Q_{\max}$ or any other bound.

\bibliography{references}